\documentclass{article}
\usepackage{iclr2027_conference,times}

\usepackage{amsmath,amssymb,amsfonts}
\DeclareMathOperator*{\argmax}{arg\,max}
\usepackage{graphicx}
\usepackage{booktabs}
\usepackage{multirow}
\usepackage{algorithm}
\usepackage{algorithmic}
\usepackage{hyperref}
\usepackage{url}
\usepackage{xcolor}
\usepackage{subcaption}
\usepackage{enumitem}
\usepackage{listings}
\usepackage[section]{placeins}
\usepackage{xspace}
\usepackage{comment}

\hypersetup{colorlinks=true, linkcolor=blue, citecolor=blue, urlcolor=blue}

\newcommand{\sys}{RLDiscover\xspace}

\definecolor{codegreen}{rgb}{0.15,0.55,0.15}
\definecolor{codegray}{rgb}{0.5,0.5,0.5}
\definecolor{backcolour}{rgb}{0.97,0.97,0.97}
\lstdefinestyle{python}{
    backgroundcolor=\color{backcolour},
    commentstyle=\color{codegreen},
    keywordstyle=\color{blue},
    basicstyle=\footnotesize\ttfamily,
    breaklines=true,
    frame=single,
    language=Python,
}

\title{\sys: LLM-Driven Co-Evolution\\of Reinforcement Learning Algorithms}

\author{%
  \parbox{\textwidth}{\centering
    Haoran Li$^1$, Zengle Ge$^2$, Xiaomin Yuan$^2$, YUI LO$^2$, Songlin Zhou$^3$,
    Jiahua Ying$^2$, Haoxin Li$^3$, \\[2pt]
    Qianhui Liu$^4$, Yuanhang Liu$^3$, Jiaqun Liu$^5$, GUOKAI CHEN$^3$,
    Mingju Chen$^6$, Ruinan Wang$^7$, \\[2pt]
    Annan Li$^2$, Jianmin Wu$^2$, Dawei Yin$^2$, Dou Shen$^2$ \\[6pt]
    {\small $^1$University of the Chinese Academy of Sciences \quad $^2$Baidu \quad $^3$Tsinghua University} \\[2pt]
    {\small $^4$Institute of Automation, Chinese Academy of Sciences \quad $^5$Peking University \quad $^6$Shanghai University \quad $^7$University of Bristol}
  }
}

\begin{document}
\AtBeginDocument{\lhead{}}

\maketitle
\begin{abstract}

LLM-guided program evolution has enabled discoveries in mathematics
and computational optimization, raising the prospect of reinforcement
learning (RL) algorithms that self-evolve to improve how agents learn.
However, realizing this prospect faces two obstacles. Joint search over coupled
algorithmic components is difficult to scale: simultaneous changes
can disrupt learning, while isolated changes overlook their
dependencies. Evaluating candidate algorithms also requires costly
training, with fitness remaining uncertain across random seeds.
We introduce \sys, a framework for the self-evolution of model-free
deep RL algorithms. Progressive Co-Evolution advances from targeted
component edits to joint evolution, while Progressive Probabilistic
Evaluation balances search breadth and evaluation fidelity through
staged training and repeated evaluation. Experiments across SAC,
PPO, and DQN on four benchmark suites show substantial improvements
in mean return, with per-family median gains of $32\%$--$84\%$
and a peak return ratio of approximately $363\times$ over a near-zero
baseline. These gains include transitions from failed learning to
successful task completion, and improvements persist when evolution
starts from stronger open-source implementations. On measured SAC
locomotion runs, evaluation uses approximately one-fifteenth the
estimated compute required to fully evaluate the same candidate pool.
Remarkably, independent searches repeatedly discover interpretable combinations
of adaptive robust losses, progress-dependent value targets, and
running statistics, with selected programs transferring to unseen
tasks. These findings point toward a broader role for self-evolution
in AI: discovering interpretable algorithms that improve how agents learn.\footnote{Code and data will be publicly released upon acceptance.}
\end{abstract}

\section{Introduction}
\label{sec:intro}

LLM-guided program evolution makes executable algorithmic logic a
direct object of search: language models propose code changes, and
evaluation feedback guides their selection~\citep{lehman2022elm}.
FunSearch demonstrated the potential of this approach for mathematical
discovery~\citep{romera2024funsearch}; AlphaEvolve extended it to
larger programs and computational optimization
tasks~\citep{novikov2025alphaevolve}; and ShinkaEvolve developed
mechanisms for more sample-efficient program
evolution~\citep{lange2025shinkaevolve}.
These advances motivate applying program evolution to the learning
procedures that govern how artificial agents improve through experience.
Reinforcement learning (RL) offers a natural target: an RL algorithm
is itself a program specifying how experience is transformed into
updates to an agent's predictions and behavior~\cite{oh2020lpg}.
We focus on model-free deep RL, encompassing value-based methods
such as DQN~\citep{mnih2015dqn}, actor-critic methods such as PPO~\citep{schulman2017proximal}, and
SAC~\citep{haarnoja2018soft}.
These algorithms specify how experience is converted into
learning targets, value estimates, and policy updates.
Their implementations therefore offer a structured space for
discovering improvements to how agents learn.

However, evolving interacting components of deep RL algorithms
faces two difficulties.
First, \emph{joint search over coupled components is difficult to scale}.
Broad randomization often yields little learning, while the same
design change can have opposite effects: increasing data reuse
can help PPO but hurt vanilla policy gradients
\citep{andrychowicz2021matters}.
These observations show that broad simultaneous changes
can disrupt learning, while optimizing choices in a fixed context
can overlook their dependencies.
Second, \emph{candidate fitness is expensive and imperfectly observed}.
In our experiments, fully evaluating 230 candidate programs would
cost approximately 1330 GPU-hours per search run
(Appendix~\ref{app:ppe}, Table~\ref{tab:ppe_stats}).\footnote{
The evaluation strategy introduced below uses approximately
87 GPU-hours for this candidate volume, a $15.3\times$ reduction
relative to the estimated full-evaluation cost.
}
Moreover, full-length training does not eliminate uncertainty caused by random seeds: 
On \texttt{mw-pick-place-v3}, the same evolved program obtains returns
of 2.9, 2887.3, and 3.9 across three seeds (Appendix~\ref{app:perseed}).
These observations expose competing demands on the evaluation budget:
exploring many programs requires inexpensive feedback, while determining
whether apparent improvements persist requires more extensive,
repeated training.

This motivates \sys, a framework that addresses these challenges through
structured proposals and staged evaluation. Its search space consists
of \textbf{five algorithmic components} within a working RL implementation.\footnote{
The interfaces cover common RL computations: observation processing,
target/advantage estimation, value-function losses, policy objectives,
and reward processing with exploration bonuses
\citep{bou2023torchrl,burda2019exploration}.
Their use is trainer-dependent.
Appendix~\ref{app:impl} gives the full interface specifications.}
\textbf{Progressive Co-Evolution (PCE)} first allocates mutations among
individual components using a UCB-inspired rule, then permits joint
revisions across components, allowing the search to build on targeted
changes while exploring their interactions.
\textbf{Progressive Probabilistic Evaluation (PPE)} allocates training
compute through increasingly demanding evaluations, reserving full
training for a small set of finalists. Together, they implement a simple
design principle: structure the search to generate useful candidates,
and increase evaluation effort as the candidate pool is narrowed.
We instantiate the framework on SAC, PPO, and DQN, covering off-policy
and on-policy learning as well as continuous and discrete control.
Our contributions are:
\begin{enumerate}
    \item \textbf{Efficient RL algorithm evolution.}
    We introduce PCE and PPE to address component dependencies
    and costly, noisy evaluations. On measured SAC locomotion
    runs, they reduce evaluation cost by approximately $15\times$
    relative to evaluating the same number of candidates
    at full fidelity.
    \item \textbf{Broad improvements across algorithms and benchmarks.}
    Across four benchmark suites, \sys improves mean return
    on all 26 tested task--algorithm pairs spanning SAC, PPO, and DQN,
    with median gains of $39.8\%$, $83.6\%$,
    and $32.1\%$, respectively. Improvements also persist when
    evolution starts from stronger open-source SAC implementations,
    with positive gains on all nine tested tasks.
    \item \textbf{Convergent discovery of interpretable mechanisms.}
    Task-specific searches repeatedly recover shared mechanisms,
    including robust critic losses in 16/17 SAC and 5/5 PPO
    champions. Cross-task evaluations further demonstrate
    transfer of evolved programs to unseen tasks, showing that program evolution recovers reusable learning mechanisms beyond hyperparameter tuning.
\end{enumerate}


\section{Related Work}
\label{sec:related}

\paragraph{Discovering RL algorithms.}
Prior work discovers learning objectives and update rules through
evolutionary optimization and meta-learning.
Evolved Policy Gradients~\citep{houthooft2018evolved} and MetaGenRL~\citep{kirsch2019improving} learn parameterized
objectives,
while LPG\citep{oh2020lpg} and DiscoRL~\citep{oh2025disco} meta-learn neural update
rules.
AutoML-Zero and Evolving RL Algorithms search programs or computation
graphs constructed from predefined operation
libraries~\citep{real2020automlzero,coreyes2021evolving}.
Eureka instead generates reward programs within an existing learning
pipeline~\citep{ma2023eureka}.
A parallel line applies the same foundation-model code-evolution paradigm
to adversarial multi-agent games, where the evaluation landscape shifts as
strategies improve and reliable assessment therefore requires evaluator
co-evolution and hierarchical evaluation~\citep{li2026beyond}.
Our focus is source-code modification across multiple components of
working RL implementations. Through fixed function interfaces,
\sys edits observation processing, targets, objectives, and reward
shaping, and studies how to coordinate these modifications under
a limited evaluation budget.

\paragraph{Multi-fidelity evaluation and search allocation.}
Successive Halving~\citep{jamieson2016successive} and Hyperband allocate~\citep{li2018hyperband} increasing resources to
promising configurations;
BOHB combines this allocation with model-guided candidate
generation~\citep{falkner2018bohb}.
PPE applies staged evaluation to evolving RL programs to balance candidate coverage against the reliability of performance estimates.
For proposal allocation, PCE uses a UCB-inspired rule to select
components for mutation~\citep{auer2002finite}, before allowing
joint program revisions.
These mechanisms adapt established allocation ideas to the coupled
search and evaluation of RL implementations.

\section{Method}
\label{sec:method}

\begin{figure}[t]
\centering
\includegraphics[width=0.78\textwidth]{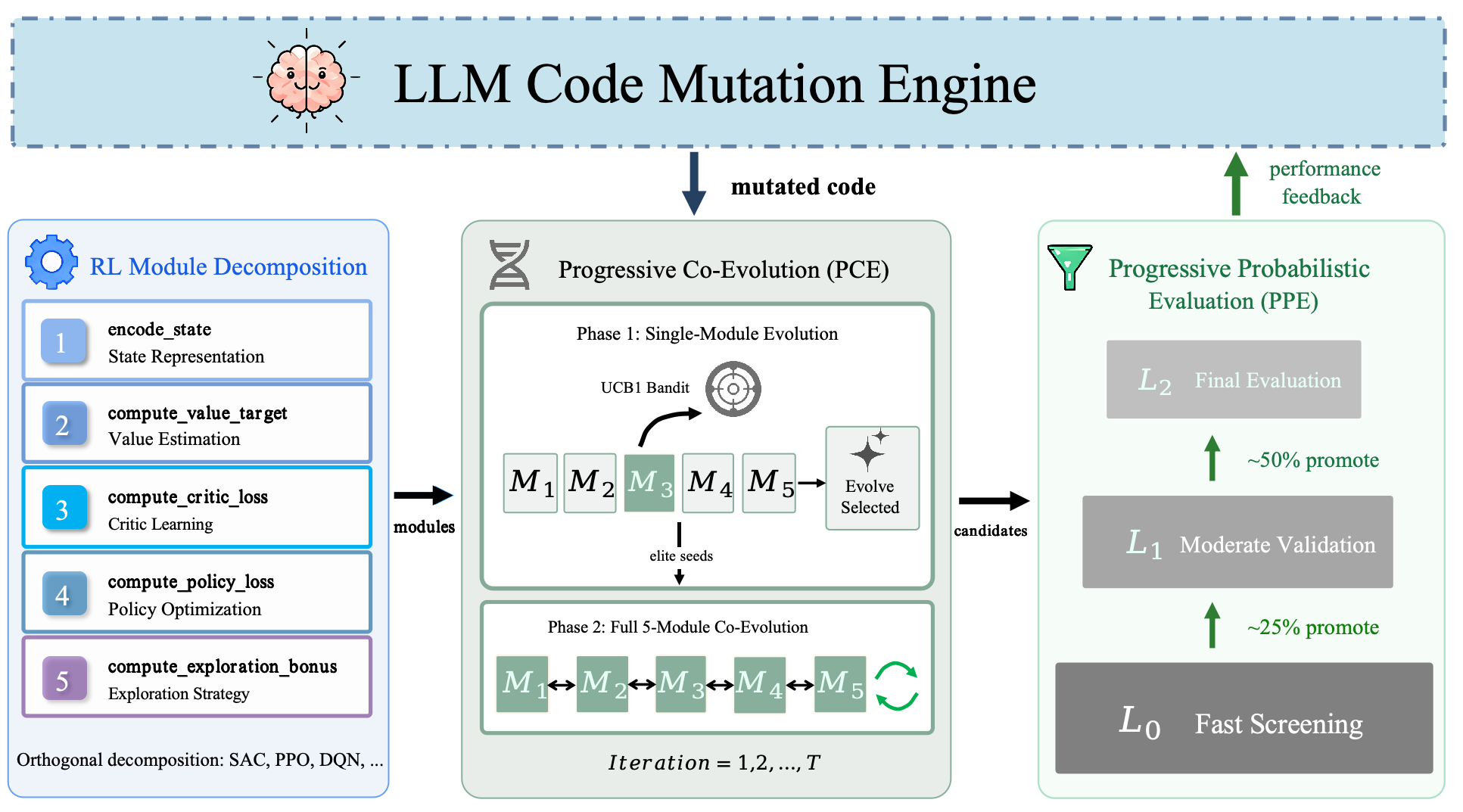}
\caption{\sys overview. \textbf{Left}: 5 algorithm components of RL algorithms (state representation, value estimation, critic learning, policy optimization, exploration). \textbf{Center}: PCE evolves components independently (Phase 1, UCB1 selection), then co-evolves them jointly (Phase 2). \textbf{Right}: PPE filters candidates through L0$\to$L1$\to$L2, so only candidates that survive cheap screening consume a full-length training run. \textbf{Top}: the LLM mutation engine generates code variants guided by performance feedback.}
\label{fig:framework}
\end{figure}

\subsection{Five-Component Decomposition}
\label{sec:components}

Given a base RL algorithm, let $\mathbf m=(m_1,\ldots,m_K)$,
with $K=5$, denote the implementations of its five algorithmic
components. Let $\mathcal A(\mathbf m)$ be the algorithm obtained
by inserting these components into the base implementation.
For a fixed task environment $e$, let $\mathcal S$ be a finite
set of training seeds and let $s$ be sampled uniformly from
$\mathcal S$. We seek to maximize
\begin{equation}
F(\mathbf m)
=
\mathbb E_{s\sim\operatorname{Unif}(\mathcal S)}
\left[
R_e\!\left(
\mathrm{Train}(\mathcal A(\mathbf m),e,s)
\right)
\right],
\end{equation}
where $\mathrm{Train}(\mathcal A,e,s)$ returns the policy obtained
by training algorithm $\mathcal A$ in environment $e$ with seed
$s$ under the full-fidelity training budget.
For a policy $\pi$, $R_e(\pi)$ denotes its mean evaluation return
in $e$, normalized using task-specific bounds.
Thus, $F(\mathbf m)$ measures the performance
of the resulting algorithm.
We write $\widehat F_\ell(\mathbf m)$ for the observed fitness
computed using the training budget.

\subsection{Progressive Co-Evolution (PCE)}
\label{sec:pce}

PCE searches for improvements in $F(\mathbf m)$ by first
restricting mutations to individual components and then
allowing coordinated changes across components
(Algorithm~\ref{alg:pce}).
This separates the search for useful local changes from
the exploration of their interactions.

\paragraph{Phase 1: bandit-guided single-component evolution.}
Each component is an arm of a UCB1 bandit~\citep{auer2002finite}. Writing $n_k$ for the number of pulls of arm $k$, $\bar{r}_k$ for its mean reward, $t$ for the total pulls so far, and $\mathcal{K}_t$ for the arms still active, the next component to mutate is
\begin{equation}
\label{eq:ucb1}
k_t = \argmax_{k \in \mathcal{K}_t} \left[ \bar{r}_k + c_t \sqrt{\frac{\ln t}{n_k}} \right],
\end{equation}
with $K{=}5$ arms and a decaying exploration coefficient $c_t = c_0/\sqrt{1 + t/K}$, $c_0{=}\sqrt{2}$ and $c_t$ shifting the bandit toward exploitation as Phase~1 proceeds.
For the selected component $k$, the LLM generates a batch
of proposed programs $\mathbf m'$, each changing only $m_k$.
These candidates are screened using $\widehat F_0(\mathbf m')$
(Section~\ref{sec:ppe}).
The arm reward is the batch's nonnegative relative fitness
gain over that component's incumbent.
After initial coverage of all components, Phase 1 freezes
components whose relative gains remain below a threshold
for consecutive batches.
It ends when all components are frozen or its budget cap
is reached, returning per-component elites and their observed gains.

\paragraph{Phase 1 $\to$ 2 handoff.}
Component champions are merged in descending order of their
Phase-1 gains. A proposed replacement $\mathbf m'$ is retained
only if $\widehat F_\ell(\mathbf m')
\ge \widehat F_\ell(\mathbf m)$ under the same validation protocol.
The resulting program initializes Phase 2. The insight library (below) is already populated from Phase~1 batches, arm-freeze comparisons, and rejected components, so Phase~2 starts from accumulated findings, negative ones included.

\paragraph{Phase 2: round-based full co-evolution.}
Phase 2 uses the remaining budget to pursue improvements
in the same objective $F(\mathbf m)$, with mutations now
allowed to change any subset of components. Each round launches a fresh population-based search seeded with the current best program. Discoveries carry across rounds via an \emph{insight library}, accumulated from Phase~1 onwards: after each round the mechanisms distinguishing that round's champions are extracted into short natural-language entries and injected into the next round's mutation prompt, so later rounds build on earlier findings.

\subsection{Progressive Probabilistic Evaluation (PPE)}
\label{sec:ppe}

PPE allocates evaluation compute to identify high-performing
programs under the objective $F(\mathbf m)$.
Let $\widehat F_\ell(\mathbf m)$ denote the observed fitness
at fidelity $\ell$. These observations differ from the
full-training objective because of training truncation and
seed-dependent variability.
PPE screens candidates through three fidelity levels:
\textbf{L0} (50K steps, one seed),
\textbf{L1} (200K steps, two seeds), and
\textbf{L2} (1M steps, three seeds).
A single L2 evaluation costs 4--10 GPU-hours in our measurements.
The resulting hierarchy supports broad candidate screening
while reserving expensive validation for a smaller pool
(Appendix~\ref{app:impl}, Table~\ref{tab:ppe_levels}).

\paragraph{Leader-referenced promotion.}
Promotion to L1 is governed by a global quota $Q$.
Let $\mathcal C_r$ denote the candidates evaluated at L0
in round $r$, ordered by decreasing $\widehat F_0$ as
$\mathbf m_{(1)},\mathbf m_{(2)},\ldots$, and let
$\mathcal P$ be the accumulated L1 pool.
Once $\mathcal P$ is nonempty, define
\begin{equation}
\mathbf p^\star
\in \arg\max_{\mathbf p\in\mathcal P}
    \widehat F_1(\mathbf p).
\end{equation}
The gate compares new candidates with the current
higher-fidelity leader on the same low-fidelity scale.
Round $r$ promotes
\begin{equation}
\label{eq:ppe_promote}
\mathcal C_r^\uparrow
=
\{\mathbf m_{(1)}\}
\;\cup\;
\left\{
\mathbf m_{(i)}:
i>1,\;
\widehat F_0(\mathbf m_{(i)})>\widehat F_0(\mathbf p^\star)
\right\},
\end{equation}
truncated in L0 rank order to the remaining quota.
The round's highest-scoring candidate therefore receives
an L1 evaluation whenever quota remains, even if no candidate
exceeds the gate.
In the final rounds, an underspent quota triggers a switch
to the pool's median L0 score as the promotion threshold.

\paragraph{Final selection.}
The five highest-scoring programs in $\mathcal P$
under $\widehat F_1$ are evaluated at L2.
The reported program is the one with the highest
$\widehat F_2$ among the audited L2 candidates.
Ranking at every level uses untruncated normalized fitness.
Before reporting, each champion is reloaded, checked for
the required component functions, and compared with the
seed program to detect silent fallbacks.
AST-hash deduplication avoids repeated evaluation of
syntactically identical programs.
Prompt construction, syntax validation, and audit details
are provided in Appendix~\ref{app:impl}.

\section{Experiments}
\label{sec:setup}
\label{sec:results}

\paragraph{Tasks.} SAC is evolved on 17 continuous control tasks---4 DMControl~\citep{tassa2018deepmind}, 5 Gymnasium~\citep{towers2024gymnasium}, 5 Meta-World~\citep{yu2019metaworld} and 3 MetaDrive~\citep{li2022metadrive}---covering locomotion, manipulation and autonomous driving, with observation dimensions 11--348 and action dimensions 2--17; PPO is evolved on 5 Gymnasium locomotion tasks and DQN on 4 discrete control tasks (CartPole, MountainCar, Acrobot, LunarLander): 26 task--algorithm pairs in total. Per-task specifications are in Appendix~\ref{app:tasks}.

\paragraph{Two kinds of seed program.} The \emph{own-init} batch seeds evolution from our own standard implementation of the algorithm and covers all 26 pairs; the \emph{SOTA-init} batch re-seeds the 9 SAC locomotion tasks that have an open-source reference---denisyarats/pytorch\_sac~\citep{yarats2019pytorch} for DMControl, CleanRL SAC~\citep{huang2022cleanrl} for Gymnasium---from that reference at unchanged capacity and hyperparameters. Each batch is compared only against the baseline it was seeded from and neither filters out any task, so each improvement is attributable to evolution rather than to a change of implementation. Seed programs, hyperparameters, the candidate budget and the L2 evaluation protocol are in Appendix~\ref{app:impl}.

\subsection{Results Across Three Algorithm Families}
\label{sec:cross_algo}

Table~\ref{tab:main_results} reports every per-task L2 cell for all three instantiations. All three are evolved by the same pipeline---same algorithm-component decomposition, same PPE funnel, same L2 protocol---differing only in the seed implementation and component calling conventions, with DQN's policy-loss slot fixed rather than searched (Table~\ref{tab:algo_component_map}). Evolution improves the baseline it was seeded from on \textbf{all 26 task--algorithm pairs}, with median gains of $+39.8\%$ on SAC, $+83.6\%$ on PPO and $+32.1\%$ on DQN, and it does so from one pipeline and one budget per task rather than per-task tuning. The gains span three orders of magnitude, from $+0.9\%$ to $\times$363.

\begin{table}[!htbp]
\centering
\footnotesize
\caption{Full per-task results across the three algorithm families, all at L2 with 3 seeds $\{0,42,123\}$ (protocol in Appendix~\ref{app:impl}; MountainCar-v0 is a 10-seed exception). Every block compares an evolved program against the baseline it was seeded from: our own SAC baseline (17 tasks), an open-source SOTA seed on all 9 locomotion tasks with an external reference (Section~\ref{sec:sota_init}; no task is filtered out), and the standard PPO / DQN implementation of our framework. The SOTA-init block re-evolves the \emph{same} 9 locomotion tasks and is not counted again in the 26. Fitness is the task-normalized return; Improvement is computed on raw returns; bold = improvement. Markers $\dagger$, $^\S$ and $^\P$ are explained in Appendix~\ref{app:family_notes}.}

\label{tab:main_results}
\setlength{\tabcolsep}{2pt}
\renewcommand{\arraystretch}{1.15}

\begin{tabular*}{\textwidth}{@{}@{\extracolsep{\fill}}llrrrr@{}}
\toprule
Task & Init & Baseline Return & Evolved Return & Fitness & Improvement \\
\midrule

\multicolumn{6}{@{}l@{}}{\makebox[0pt][l]{%
\parbox[t]{\textwidth}{\raggedright
\textit{SAC, own-init (17 tasks): evolved vs.\ our own SAC baseline; evolved $^\S$ --- median $+39.8\%$}}}}\\
\midrule

dm-cheetah-run
& Ours & 302.3{\scriptsize$\,\pm$37.3}
& 646.9{\scriptsize$\,\pm$67.4}
& 0.647 & \textbf{+114.0\%} \\
dm-walker-walk
& Ours & 679.6{\scriptsize$\,\pm$41.0}
& 971.0{\scriptsize$\,\pm$3.3}\smash{$^\dagger$}
& 0.971 & \textbf{+42.9\%} \\
dm-walker-run
& Ours & 487.7{\scriptsize$\,\pm$56.3}
& 585.7{\scriptsize$\,\pm$7.1}
& 0.586 & \textbf{+20.1\%} \\
dm-quadruped-walk
& Ours & 931.6{\scriptsize$\,\pm$12.5}
& 939.8{\scriptsize$\,\pm$4.1}\smash{$^\dagger$}
& 0.940 & \textbf{+0.9\%} \\
gym-HalfCheetah-v5
& Ours & 5093.6{\scriptsize$\,\pm$209.3}
& 9053.7{\scriptsize$\,\pm$1554.1}
& 0.604 & \textbf{+77.7\%} \\
gym-Hopper-v5
& Ours & 2475.0{\scriptsize$\,\pm$613.7}
& 2780.7{\scriptsize$\,\pm$861.6}\smash{$^\dagger$}
& 0.695 & \textbf{+12.4\%} \\
gym-Walker2d-v5
& Ours & 3883.1{\scriptsize$\,\pm$218.3}
& 5429.5{\scriptsize$\,\pm$485.3}\smash{$^\dagger$}
& 0.905 & \textbf{+39.8\%} \\
gym-Humanoid-v5
& Ours & 4998.8{\scriptsize$\,\pm$205.5}
& 5281.2{\scriptsize$\,\pm$152.7}\smash{$^\dagger$}
& 0.755 & \textbf{+5.6\%} \\
gym-Ant-v5
& Ours & 3732.8{\scriptsize$\,\pm$516.1}
& 3785.9{\scriptsize$\,\pm$821.9}
& 0.473 & \textbf{+1.4\%} \\
mw-push-v3
& Ours & 5.0{\scriptsize$\,\pm$0.4}
& 1825.1{\scriptsize$\,\pm$1875.4}
& 0.406 & \textbf{$\times$363} \\
mw-drawer-open-v3
& Ours & 3177.0{\scriptsize$\,\pm$1190.9}
& 4621.3{\scriptsize$\,\pm$54.9}
& 1.000 & \textbf{+45.5\%} \\
mw-door-open-v3
& Ours & 2906.0{\scriptsize$\,\pm$889.8}
& 3099.9{\scriptsize$\,\pm$1923.0}
& 0.689 & \textbf{+6.7\%} \\
mw-peg-insert-side-v3
& Ours & 1151.9{\scriptsize$\,\pm$1607.1}
& 3075.6{\scriptsize$\,\pm$2172.8}
& 0.683 & \textbf{+167.0\%} \\
mw-pick-place-v3
& Ours & 6.8{\scriptsize$\,\pm$2.4}
& 964.7{\scriptsize$\,\pm$1359.5}
& 0.214 & \textbf{$\times$141} \\
md-validation
& Ours & 147.8{\scriptsize$\,\pm$85.8}
& 323.2{\scriptsize$\,\pm$150.3}
& 0.923 & \textbf{+118.7\%} \\
md-10env
& Ours & 229.6{\scriptsize$\,\pm$57.6}
& 232.7{\scriptsize$\,\pm$27.1}
& 0.665 & \textbf{+1.4\%} \\
md-100envs
& Ours & 237.1{\scriptsize$\,\pm$60.1}
& 282.2{\scriptsize$\,\pm$22.4}
& 0.806 & \textbf{+19.0\%} \\

\midrule
\multicolumn{6}{@{}l@{}}{\makebox[0pt][l]{%
\parbox[t]{\textwidth}{\raggedright
\textit{SAC, SOTA-init (9 tasks, not counted again): evolved vs.\ denisyarats / CleanRL seed; evolved $^\S$ --- median $+2.6\%$}}}}\\
\midrule

dm-cheetah-run
& denisyarats & 713.3{\scriptsize$\,\pm$140.1}
& 855.8{\scriptsize$\,\pm$33.7}
& 0.856 & \textbf{+20.0\%} \\
dm-walker-walk
& denisyarats & 967.5{\scriptsize$\,\pm$1.3}
& 972.8{\scriptsize$\,\pm$3.6}
& 0.973 & \textbf{+0.5\%} \\
dm-walker-run
& denisyarats & 753.0{\scriptsize$\,\pm$7.7}
& 772.5{\scriptsize$\,\pm$4.2}
& 0.773 & \textbf{+2.6\%} \\
dm-quadruped-walk
& denisyarats & 924.9{\scriptsize$\,\pm$14.1}
& 940.8{\scriptsize$\,\pm$17.3}
& 0.941 & \textbf{+1.7\%} \\
gym-HalfCheetah-v5
& CleanRL & 11002.4{\scriptsize$\,\pm$91.5}
& 11064.9{\scriptsize$\,\pm$240.2}
& 0.738 & \textbf{+0.6\%} \\
gym-Hopper-v5
& CleanRL & 2360.8{\scriptsize$\,\pm$624.8}
& 2832.5{\scriptsize$\,\pm$520.3}
& 0.708 & \textbf{+20.0\%} \\
gym-Walker2d-v5
& CleanRL & 4672.3{\scriptsize$\,\pm$354.0}
& 4689.8{\scriptsize$\,\pm$208.6}
& 0.782 & \textbf{+0.4\%} \\
gym-Ant-v5
& CleanRL & 4074.7{\scriptsize$\,\pm$793.5}
& 5188.8{\scriptsize$\,\pm$242.9}
& 0.649 & \textbf{+27.3\%} \\
gym-Humanoid-v5
& CleanRL & 5165.5{\scriptsize$\,\pm$414.1}
& 5643.7{\scriptsize$\,\pm$297.3}
& 0.806 & \textbf{+9.3\%} \\

\midrule
\multicolumn{6}{@{}l@{}}{\makebox[0pt][l]{%
\parbox[t]{\textwidth}{\raggedright
\textit{PPO (5 tasks): evolved vs.\ standard PPO baseline; evolved $^\S$ --- median $+83.6\%$}}}}\\
\midrule

gym-Hopper-v5
& Ours & 1920.0{\scriptsize$\,\pm$289.9}
& 3525.5{\scriptsize$\,\pm$57.8}
& 0.881 & \textbf{+83.6\%} \\
gym-HalfCheetah-v5
& Ours & 1724.4{\scriptsize$\,\pm$39.6}
& 4233.6{\scriptsize$\,\pm$349.8}
& 0.282 & \textbf{+145.5\%} \\
gym-Ant-v5
& Ours & 776.0{\scriptsize$\,\pm$76.3}
& 2789.0{\scriptsize$\,\pm$445.0}
& 0.349 & \textbf{+259.4\%} \\
gym-Walker2d-v5
& Ours & 4023.4{\scriptsize$\,\pm$655.0}
& 4440.2{\scriptsize$\,\pm$478.3}
& 0.740 & \textbf{+10.4\%} \\
gym-Humanoid-v5
& Ours & 606.6{\scriptsize$\,\pm$55.1}
& 809.1{\scriptsize$\,\pm$91.7}
& 0.116 & \textbf{+33.4\%} \\

\midrule
\multicolumn{6}{@{}l@{}}{\makebox[0pt][l]{%
\parbox[t]{\textwidth}{\raggedright
\textit{DQN (4 tasks): evolved vs.\ standard DQN baseline; evolved $^\S$ --- median $+32.1\%$}}}}\\
\midrule

CartPole-v1
& Ours & 330.2{\scriptsize$\,\pm$18.1}
& 451.2{\scriptsize$\,\pm$69.0}
& 0.902 & \textbf{+36.6\%} \\
MountainCar-v0\smash{$^\P$}
& Ours & $-$200.0{\scriptsize$\,\pm$0.0}
& $-$144.9{\scriptsize$\,\pm$29.2}
& 0.937 & \textbf{+27.5\%} \\
LunarLander-v3
& Ours & 66.9{\scriptsize$\,\pm$87.0}
& 174.6{\scriptsize$\,\pm$24.2}
& 0.582 & \textbf{+161.0\%} \\
Acrobot-v1
& Ours & $-$91.2{\scriptsize$\,\pm$5.3}
& $-$87.0{\scriptsize$\,\pm$6.6}
& 0.983 & \textbf{+4.7\%} \\

\midrule
\multicolumn{6}{@{}l@{}}{\textit{26/26 audited improvements across the three families (17 SAC own-init, 5 PPO, 4 DQN)}} \\
\bottomrule
\end{tabular*}
\end{table}

\paragraph{Search can reorder the algorithms themselves.} PPO shows the largest median gain of the three families ($+83.6\%$), consistent with PPO's seed program having the lowest absolute starting point of the three families---more room to improve, larger gains. On gym-Hopper-v5, evolved PPO ($3525.5\pm57.8$) ends above our SAC champion on that task ($2780.7\pm861.6$), with a fifth of the seed spread---the weaker family evolved past the stronger one. Which family to start from is itself a search outcome, not something fixed in advance. The comparison holds because both numbers are raw returns on the same task under the same evaluation protocol; \emph{fitness} is normalised per family and is not comparable across them, and neither is comparable to published numbers. Per-family qualifications, the $\dagger$ comparisons against external references and the normalization artifact behind MountainCar's high baseline fitness are in Appendix~\ref{app:family_notes}.

\paragraph{Programs are diagnosable when search stalls.} gym-Ant-v5 has the smallest own-init gain in the suite ($+1.4\%$). Reading the champion as code, rather than as a score, reveals why: three of the components are unchanged from the seed, and Ant's champion is the only one in the suite that kept plain MSE as the critic loss---the one change that virtually every other champion made (Appendix~\ref{app:components}). This points to insufficient search, not a task ceiling: the search did not explore enough from this starting point. The fix the diagnosis suggests is a stronger seed; re-seeding from CleanRL turns Ant's $+1.4\%$ into the largest gain in the SOTA-init block ($+27.3\%$). A fitness score cannot tell a hard task from an undersearched one; the program can.

\subsection{Evolution from Stronger SOTA Baselines}
\label{sec:sota_init}
\label{sec:stronger_baseline}

The \emph{SOTA-init} block of Table~\ref{tab:main_results} addresses a natural concern: are the gains partly due to a weak starting implementation? The external baselines are indeed substantially stronger on several tasks (CleanRL reaches 11002.4 vs.\ our 5093.6 on HalfCheetah). Re-seeding from them, evolution improves on the SOTA starting point on \textbf{all 9 of 9} tasks, never degrades a strong seed, and finds double-digit gains on 3 (Ant $+27.3\%$, Hopper $+20.0\%$, cheetah-run $+20.0\%$).

The more informative quantity is what happens to the \emph{gap} between the two batches, since that is what the concern is actually about. At the seed program the two implementations differ by a mean of $23.4\%$ in return; after evolution the mean gap is $13.1\%$, narrowing on 7 of the 9 tasks---on dm-walker-walk almost entirely ($29.8\%\to0.2\%$). Evolution recovers roughly half of an implementation-quality gap it did not create, neither closing nor widening it. The two tasks where the gap instead \emph{widens} are Humanoid and Ant, the latter being the same cell identified above as the one where own-init search failed: a stronger seed helps most exactly where the search itself did least.

Narrowing is not the same as converging, however, and the head-to-head outcome is a separate question. On the nine locomotion tasks---the only ones both batches were evolved on with 3 seeds---the SOTA-init champion finishes ahead on 8 of 9, losing only on Walker2d; two of those eight are ties of $0.2\%$ or less, so the honest reading is six clear wins, one loss, and two ties. The gap narrows on 7 of 9 tasks while the ranking survives on only 4, and where it flips the per-task search gain rather than the seed decides it: Walker2d is a genuine overtake, Ant the mirror image. A large seed advantage is durable and a small one is not---all four preserved rankings sit on seed gaps of at least $30\%$. Per-task seed gaps, final gaps, and each batch's gain over its seed are in Appendix~\ref{app:sota_gaps}. The core innovations (Huber loss, $Q$-annealing) emerge in both batches---what we would expect of improvements to the algorithm itself rather than artifacts of a particular starting implementation.

\subsection{Discovered Algorithmic Innovations}
\label{sec:innovations}

\begin{figure}[t]
\centering
\begin{subfigure}[t]{0.40\textwidth}
\centering
\includegraphics[width=\textwidth]{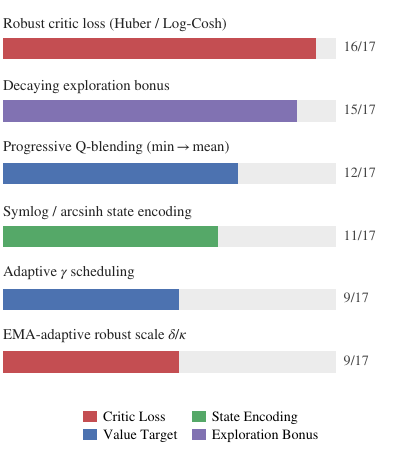}
\caption{Consensus innovations (17 champions)}
\label{fig:consensus}
\end{subfigure}
\hfill
\begin{subfigure}[t]{0.40\textwidth}
\centering
\includegraphics[width=\textwidth]{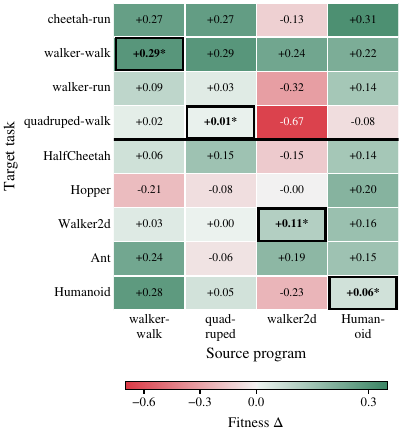}
\caption{Cross-task transfer of 4 champions}
\label{fig:cross_task}
\end{subfigure}
\caption{Evolution finds algorithmic improvements rather than per-task overfits. \textbf{(a)} Frequency of independently discovered innovations across all 17 champions, coloured by the component each modifies. \textbf{(b)} Fitness $\Delta$ from running a source champion (column) on a target task (row); green = improvement, red = regression; boxed cells marked $*$ are the source task itself. Criteria, the full $17\times6$ matrix and per-task values: Appendices~\ref{app:innovations} and~\ref{app:generalization}.}
\label{fig:innovations}
\end{figure}

Independent evolution repeatedly converges on a small set of core modifications (Figure~\ref{fig:consensus}), and the programs carrying them transfer to tasks they were never evolved on (Figure~\ref{fig:cross_task}); we take each task's best champion across both seed batches, covering all 17 tasks. Reading the PPO and DQN champions the same way (Appendix~\ref{app:components_xalgo}) isolates exactly one innovation shared by all three families---replacing MSE with a robust regression critic loss, found in 5/5 PPO and 3/4 DQN champions alongside SAC's 16/17, i.e.\ 8 of the 9 cross-algorithm champions under two different trainers. Each family's \emph{other} innovations sit in machinery specific to that family: GAE-$\lambda$ scheduling and clipped-surrogate/KL control in 5/5 PPO champions, $Q$-annealing in SAC, which PPO cannot express because it has no twin critics. DQN is the honest exception: only three of its five slots are live under our trainer (the policy slot is fixed and the value target reduces to the plain $\max_a Q'$ backup), so its champions concentrate on encoding and exploration by construction rather than by discovery, with added critic-loss changes in three (Appendix~\ref{app:components_xalgo}).

\textbf{Robust critic loss.} addresses bootstrapping noise: MSE amplifies large TD errors quadratically while robust losses cap gradients. All but one champion adopt one (Huber, smooth-L1, Log-Cosh, or Charbonnier) and 9 adapt the transition scale via an EMA of running error statistics. \textbf{Progressive Q-blending} replaces the fixed pessimistic $\min(Q_1, Q_2)$ inherited from clipped double-Q learning~\citep{fujimoto2018addressing} with a progress-annealed blend of $\min$ and mean---a schedule never previously proposed for SAC---with the annealing direction splitting by domain. \textbf{Symlog encoding} compresses observations via $\text{sign}(x)\cdot\log(1{+}|x|)$; a similar transform was independently proposed in DreamerV3~\citep{hafner2023dreamerv3}. These modifications transfer (Figure~\ref{fig:cross_task}): of four champions re-evaluated on 9 locomotion targets, three improve 6--7 of their 8 non-source tasks (mean fitness $\Delta$ $+0.083$ to $+0.157$), while the walker2d program---which relies on hard-coded step thresholds---wins only 2/8 ($-0.134$; Appendices~\ref{app:generalization},~\ref{app:innovations}).

\paragraph{Cross-domain generality.} In Meta-World, push and pick-place achieve order-of-magnitude gains from near-zero baselines, demonstrating that evolution can fundamentally restructure failing training procedures. In MetaDrive, all 3 driving scenarios improve, including 100-envs large-scale generalization (+19.0\%). The same evolution pipeline and evaluation protocol work across locomotion, robotic manipulation, and autonomous driving without domain-specific adjustments.

\paragraph{Interpreting the discovered mechanisms.} Progressive Q-blending's annealing direction splits by domain: locomotion champions start optimistic and anneal toward $\min$, while manipulation and driving champions \emph{relax} pessimism as training stabilizes. Symlog encoding prevents gradient saturation in high-dimensional tasks; MSE amplifies large TD errors quadratically while robust losses cap gradients at the transition scale, acting as built-in gradient clipping.

\paragraph{Principles for generalizable programs.} Comparing a well-generalizing program (quadruped, 6/8 wins) with the worst (walker2d, 2/8 wins), we identify three principles for transferable RL modifications. \textbf{(1) Relative scheduling}: use progress ratio $t/T$ rather than absolute step counts for annealing---the quadruped program's cosine-annealed schedules adapt to any training length. \textbf{(2) Stateless encoding}: pointwise transformations (arcsinh, symlog) transfer perfectly; running statistics do not. \textbf{(3) Adaptive thresholds}: parameters derived from training dynamics (EMA of TD errors, as in the walker-walk program) auto-calibrate to each task, while hard-coded constants may be inappropriate for different reward scales.

\section{Ablation Studies}
\label{sec:ablation}

We ablate PCE+PPE with 3 groups of 8 tasks\footnote{
The detailed configurations are:
(1)
SAC on \texttt{dm-cheetah-run},
\texttt{gym-HalfCheetah-v5},
\texttt{md-100envs}, and
\texttt{mw-drawer-open-v3};
(2)
The four SAC pairs in I,
PPO on \texttt{gym-Hopper-v5} and
\texttt{gym-HalfCheetah-v5};
and DQN on \texttt{LunarLander-v3} and
\texttt{MountainCar-v0};
(3)
The four SAC pairs in I,
PPO on \texttt{gym-Hopper-v5};
and DQN on \texttt{LunarLander-v3}.
} under the matched-budget protocol of Appendix~\ref{app:impl}. We compare the full method against five ablations. Two remove one PCE component at a time and keep everything else: \emph{w/o component decomposition} keeps Phase~1 and the funnel but makes every mutation rewrite the whole program, and \emph{w/o Phase~1} keeps the decomposition and the funnel but skips the bandit stage. The remaining three---\emph{L0}, \emph{L1} and \emph{L2}---ask what a fixed evaluation fidelity is worth, and to answer that they run the search itself in its simplest form: the whole program is mutated as one block, Phase~1 is skipped, and every candidate is scored at that single level, with the compute-aligned budget the level allows. They therefore vary \emph{three} factors against the full method at once (funnel, decomposition, Phase~1), and a gap between the full method and any of them cannot be attributed to the funnel alone. We read them as bounding what a single fidelity level can achieve rather than as isolating PPE.

\begin{table}[!htbp]
\centering
\footnotesize
\caption{Component ablation across the three algorithm families: audited L2 champion fitness, mean $\pm$ cross-task std over \textbf{3 independent evolution runs} per cell ($8$ pairs $\times$ 6 conditions $\times$ 3 runs $=144$ runs); per-task matrix in Table~\ref{tab:ablation_full}, per-cell evolution spreads in Appendix~\ref{app:ablation_spread}. The $\pm$ is the standard deviation \emph{across tasks} within each row.}
\label{tab:ablation_main}
\setlength{\tabcolsep}{1pt}
\begin{tabular*}{\textwidth}{@{\extracolsep{\fill}}lcccccc}
\toprule
 & & \multicolumn{2}{c}{Ablating PCE} & \multicolumn{3}{c}{Single-fidelity PPE} \\
\cmidrule(lr){3-4}\cmidrule(lr){5-7}
 & Full & w/o decomp. & w/o Phase~1 & L0 & L2 & L1 \\
\midrule
(1) & \textbf{0.784}{\scriptsize$\,\pm$0.09} & 0.762{\scriptsize$\,\pm$0.13} & 0.758{\scriptsize$\,\pm$0.14} & 0.728{\scriptsize$\,\pm$0.17} & 0.700{\scriptsize$\,\pm$0.08} & 0.691{\scriptsize$\,\pm$0.05} \\
\addlinespace[2pt]
(2) & \textbf{0.731}{\scriptsize$\,\pm$0.23} & 0.687{\scriptsize$\,\pm$0.25} & 0.704{\scriptsize$\,\pm$0.24} & 0.687{\scriptsize$\,\pm$0.25} & 0.700{\scriptsize$\,\pm$0.21} & 0.694{\scriptsize$\,\pm$0.22} \\
\addlinespace[2pt]
(3) & \textbf{0.789}{\scriptsize$\,\pm$0.08} & 0.731{\scriptsize$\,\pm$0.17} & 0.748{\scriptsize$\,\pm$0.13} & 0.727{\scriptsize$\,\pm$0.15} & 0.728{\scriptsize$\,\pm$0.10} & 0.736{\scriptsize$\,\pm$0.08} \\
\bottomrule
\end{tabular*}
\end{table}

\paragraph{The full method leads every ablation on the pooled mean.} It leads all five by $0.026$--$0.043$ normalised fitness over the 8 pairs and by $0.041$--$0.062$ over the 6 non-degenerate pairs, and the sign of the paired $\Delta$ is negative for every ablation in both groupings. The per-pair spread is $1.3$--$4\times$ the gap being measured, however, so the consistent ordering rests on the sign pattern rather than on any single comparison, and we read the pooled numbers as directional.

\paragraph{Stratifying by family shows why, and recovers the SAC result.} The pooled null averages over families that disagree. Within SAC the funnel effect is large and ordered: on the four-task mean the full method leads all three single-level variants by 7.1\% (L0), 10.7\% (L2) and 11.8\% (L1), with distinct failure modes---L0 searches the widest but ranks on 50K-step single-seed returns that favour fast starters not all of which survive to 1M steps; L2 ranks on the true objective but at $\sim$5.8 GPU-hours per evaluation affords only 13 candidates, degenerating into local refinement of a handful of programs; L1 inherits neither L0's breadth nor L2's fidelity. On PPO and DQN the directions are inconsistent or reversed---L2 beats the full method on \emph{both} PPO tasks and wins 4 of the 8 pairs, two of them starred. Since the funnel's thresholds were calibrated on SAC, we claim the funnel advantage for SAC only.

\paragraph{Two statements hold in every family.} \textbf{(a)} Single-level selection at the cheapest fidelity ($L_0$, the simplest search configuration) is the only ablation trailing the full method in every family. \textbf{(b)} Compute-aligned L2 evaluation is the closest competitor pooled across families, reaching comparable fitness by spending the whole budget on 13 fully-trained candidates and forgoing the funnel's wider search and early-stopping speedup. What both bound is the same trade-off: across the three configurations the search gets wider as the fidelity gets cheaper and the winner sits in between. Because the three single-level conditions also drop the decomposition and Phase~1, the full method's margin is evidence for the \emph{combination}, and the defensible claim about the funnel specifically is \emph{efficiency at equal quality} rather than demonstrably higher fitness.

\paragraph{Selection fidelity pays off where the search problem is hard.}
Separating six configurations requires headroom: if the baseline
already saturates a cell, all six cluster on the same score
and the ordering becomes noise.
The four SAC pairs split by baseline strength
(Appendix~\ref{app:ablation_spread}, Table~\ref{tab:ablation_spread}).
Where the initial solution is weak
(on dm-cheetah-run and gym-HalfCheetah-v5), even the \emph{worst}
ablated condition beats it by $+0.25$ and $+0.31$,
and the full method ranks first of six on both.
Where the initial solution already sits near the top of the range
(on md-100envs and mw-drawer-open-v3), the worst condition clears
it by only $+0.051$ and $+0.012$, the full method drops
to fourth and second, and on md-100envs the winner is
L0, the \emph{cheapest} condition.
These results suggest that selection fidelity matters most when the
baseline leaves substantial room for improvement; benchmarks dominated
by near-saturated baselines may therefore underestimate it.


\section{Conclusion}
\label{sec:conclusion}

\sys advances automated RL algorithm evolution by searching executable
learning rules under a limited evaluation budget. Progressive
Co-Evolution coordinates edits to interacting components, while
Progressive Probabilistic Evaluation concentrates training compute
on promising candidates. Across SAC, PPO, and DQN, the evolved
programs improve learning performance, including successful task
completion where the starting implementation fails. The search
uncovers interpretable combinations of adaptive robust losses,
progress-dependent value targets, and stateful schedules, whose
recurrence across independent searches and observed cross-task
transfer provide evidence of reusable algorithmic structure.
By connecting LLM-generated code with training-based selection,
\sys offers the community both a practical discovery framework
and executable mechanisms to inspect, test, and adapt---evidence
that program evolution can expand what agents learn and generate
concrete insights for designing better RL algorithms.

\subsection*{AI use statement}

Large language models are the central experimental apparatus of this work, not an authoring aid. All candidate programs reported here---every evolved component implementation, including all champion programs analysed in Section~\ref{sec:innovations} and Appendix~\ref{app:innovations}---were generated by an LLM acting as the mutation operator inside \sys, under the prompting protocol and feedback format described in Section~\ref{sec:method} and Appendix~\ref{app:impl}. The identity of the mutation model is withheld for anonymity and will be stated in the camera-ready version together with decoding settings and the exact prompt templates; the generated programs themselves will be released with the code. Every program that enters a reported result was executed and scored by the automated evaluation pipeline, and all champion programs were additionally read and audited by the authors for degenerate behaviour (reward hacking, evaluation-protocol violation, or hyperparameter changes outside the component interface); the audit criteria and outcomes are given in Appendix~\ref{app:innovations}.

Separately, the authors used an LLM assistant for routine writing support---copy-editing, LaTeX table formatting, and plotting-script boilerplate. The authors verified every number in the paper against the raw experiment logs and take full responsibility for the content.

\subsection*{Reproducibility statement}

Section~\ref{sec:setup} specifies the benchmarks, training budgets, seed sets, and the L2 evaluation protocol used for every reported number; Appendix~\ref{app:impl} gives the component interfaces, the PCE schedule, and the PPE promotion rule (Eq.~\ref{eq:ppe_promote}). Per-seed results for all 17 SAC own-init cells and all 9 PPO/DQN cells are reported in Appendix~\ref{app:perseed} rather than summarised; the 9 SOTA-init cells are covered by the per-cell $\pm$ in Table~\ref{tab:main_results} whose provenance is stated in that caption, and Section~\ref{sec:limitations} states explicitly where our sample size ($n{=}3$ per cell) supports only qualitative claims. External SOTA reference values are re-run by us under our own evaluation protocol rather than copied from the cited papers. The evolution framework, the evolved champion programs, the evaluation harness, and the per-cell result data will be released publicly upon acceptance of the paper.


\bibliographystyle{iclr2027_conference}
\bibliography{references}

@article{romera2024funsearch,
  title={Mathematical Discoveries from Program Search with Large Language Models},
  author={Romera-Paredes, Bernardino and Barekatain, Mohammadamin and Novikov, Alexander and Balog, Matej and Kumar, M Pawan and Dupont, Emilien and Ruiz, Francisco J R and Ellenberg, Jordan S and Wang, Pengming and Fawzi, Omar and Kohli, Pushmeet and Fawzi, Alhussein},
  journal={Nature},
  volume={625},
  number={7995},
  pages={468--475},
  year={2024},
  publisher={Nature Publishing Group}
}

@article{novikov2025alphaevolve,
  title={{AlphaEvolve}: A Coding Agent for Scientific and Algorithmic Discovery},
  author={Novikov, Alexander and V{\~u}, Ng{\^a}n and Eisenberger, Marvin and Dupont, Emilien and Huang, Po-Sen and Wagner, Adam Zsolt and Shirobokov, Sergey and Kozlovskii, Borislav and Ruiz, Francisco J R and Mehrabian, Abbas and Kumar, M Pawan and See, Abigail and Chaudhuri, Swarat and Holland, George and Davies, Alex and Nowozin, Sebastian and Kohli, Pushmeet and Balog, Matej},
  journal={arXiv preprint arXiv:2506.13131},
  year={2025}
}

@article{lange2025shinkaevolve,
  title={{ShinkaEvolve}: Towards Open-Ended and Sample-Efficient Program Evolution},
  author={Lange, Robert Tjarko and Imajuku, Yuki and Cetin, Edoardo},
  journal={arXiv preprint arXiv:2509.19349},
  year={2025}
}

@inproceedings{ma2023eureka,
  title={{Eureka}: Human-Level Reward Design via Coding Large Language Models},
  author={Ma, Yecheng Jason and Liang, William and Wang, Guanzhi and Huang, De-An and Bastani, Osbert and Jayaraman, Dinesh and Zhu, Yuke and Fan, Linxi and Anandkumar, Anima},
  booktitle={International Conference on Learning Representations},
  year={2024}
}

@inproceedings{houthooft2018evolved,
  title={Evolved Policy Gradients},
  author={Houthooft, Rein and Chen, Richard Y and Isola, Phillip and Stadie, Bradly C and Wolski, Filip and Ho, Jonathan and Abbeel, Pieter},
  booktitle={Advances in Neural Information Processing Systems},
  volume={31},
  year={2018}
}

@inproceedings{kirsch2019improving,
  title={Improving Generalization in Meta Reinforcement Learning using Learned Objectives},
  author={Kirsch, Louis and van Steenkiste, Sjoerd and Schmidhuber, J{\"u}rgen},
  booktitle={International Conference on Learning Representations},
  year={2020}
}

@article{oh2025disco,
  title={Discovering State-of-the-Art Reinforcement Learning Algorithms},
  author={Oh, Junhyuk and Farquhar, Gregory and Kemaev, Iurii and Calian, Dan A and Hessel, Matteo and Zintgraf, Luisa and Singh, Satinder and van Hasselt, Hado and Silver, David},
  journal={Nature},
  volume={648},
  number={8093},
  pages={312--319},
  year={2025},
  publisher={Nature Publishing Group}
}

@inproceedings{haarnoja2018soft,
  title={Soft Actor-Critic: Off-Policy Maximum Entropy Deep Reinforcement Learning with a Stochastic Actor},
  author={Haarnoja, Tuomas and Zhou, Aurick and Abbeel, Pieter and Levine, Sergey},
  booktitle={International Conference on Machine Learning},
  pages={1861--1870},
  year={2018},
  organization={PMLR}
}

@article{hafner2023dreamerv3,
  title={Mastering Diverse Domains through World Models},
  author={Hafner, Danijar and Pasukonis, Jurgis and Ba, Jimmy and Lillicrap, Timothy},
  journal={arXiv preprint arXiv:2301.04104},
  year={2023}
}

@article{schulman2017proximal,
  title={Proximal Policy Optimization Algorithms},
  author={Schulman, John and Wolski, Filip and Dhariwal, Prafulla and Radford, Alec and Klimov, Oleg},
  journal={arXiv preprint arXiv:1707.06347},
  year={2017}
}

@article{mnih2015dqn,
  title={Human-level Control through Deep Reinforcement Learning},
  author={Mnih, Volodymyr and Kavukcuoglu, Koray and Silver, David and Rusu, Andrei A and Veness, Joel and Bellemare, Marc G and Graves, Alex and Riedmiller, Martin and Fidjeland, Andreas K and Ostrovski, Georg and others},
  journal={Nature},
  volume={518},
  pages={529--533},
  year={2015}
}

@inproceedings{yu2019metaworld,
  title={{Meta-World}: A Benchmark and Evaluation for Multi-Task and Meta Reinforcement Learning},
  author={Yu, Tianhe and Quillen, Deirdre and He, Zhanpeng and Julian, Ryan and Hausman, Karol and Finn, Chelsea and Levine, Sergey},
  booktitle={Conference on Robot Learning},
  pages={1094--1100},
  year={2020},
  organization={PMLR}
}

@article{li2022metadrive,
  title={{MetaDrive}: Composing Diverse Driving Scenarios for Generalizable Reinforcement Learning},
  author={Li, Quanyi and Peng, Zhenghao and Feng, Lan and Zhang, Qihang and Xue, Zhenghai and Zhou, Bolei},
  journal={IEEE Transactions on Pattern Analysis and Machine Intelligence},
  volume={45},
  number={3},
  pages={3461--3475},
  year={2023},
  publisher={IEEE}
}

@misc{yarats2019pytorch,
  title={Soft Actor-Critic ({SAC}) Implementation in {PyTorch}},
  author={Yarats, Denis and Kostrikov, Ilya},
  year={2020},
  publisher={GitHub},
  howpublished={\url{https://github.com/denisyarats/pytorch_sac}}
}

@article{huang2022cleanrl,
  title={{CleanRL}: High-quality Single-file Implementations of Deep Reinforcement Learning Algorithms},
  author={Huang, Shengyi and Dossa, Rousslan Fernand Julien and Ye, Chang and Braga, Jeff and Chakraborty, Dipam and Mehta, Kinal and Ara{\'u}jo, Jo{\~a}o G M},
  journal={Journal of Machine Learning Research},
  volume={23},
  number={274},
  pages={1--18},
  year={2022}
}

@article{auer2002finite,
  title={Finite-time Analysis of the Multiarmed Bandit Problem},
  author={Auer, Peter and Cesa-Bianchi, Nicol{\`o} and Fischer, Paul},
  journal={Machine Learning},
  volume={47},
  number={2},
  pages={235--256},
  year={2002},
  publisher={Springer}
}

@article{tassa2018deepmind,
  title={DeepMind Control Suite},
  author={Tassa, Yuval and Doron, Yotam and Muldal, Alistair and Erez, Tom and Li, Yazhe and de Las Casas, Diego and Budden, David and Abdolmaleki, Abbas and Merel, Josh and Lefrancq, Andrew and Lillicrap, Timothy and Riedmiller, Martin},
  journal={arXiv preprint arXiv:1801.00690},
  year={2018}
}

@article{towers2024gymnasium,
  title={Gymnasium: A Standard Interface for Reinforcement Learning Environments},
  author={Towers, Mark and Kwiatkowski, Ariel and Terry, Jordan and Balis, John U. and De Cola, Gianluca and Deleu, Tristan and Goul{\~a}o, Manuel and Kallinteris, Andreas and Krimmel, Markus and KG, Arjun and Perez-Vicente, Rodrigo and Pierr{\'e}, Andrea and Schulhoff, Sander and Tai, Jun Jet and Tan, Hannah and Younis, Omar G.},
  journal={arXiv preprint arXiv:2407.17032},
  year={2024}
}

@inproceedings{fujimoto2018addressing,
  title={Addressing Function Approximation Error in Actor-Critic Methods},
  author={Fujimoto, Scott and van Hoof, Herke and Meger, David},
  booktitle={International Conference on Machine Learning},
  year={2018}
}

@article{li2026beyond,
  title={Beyond Static Evaluation: Co-Evolutionary Mechanisms for {LLM}-Driven Strategy Evolution in Adversarial Games},
  author={Anonymous},
  journal={arXiv preprint arXiv:2606.10389},
  year={2026}
}

@inproceedings{jamieson2016successive,
  title={Non-stochastic Best Arm Identification and Hyperparameter Optimization},
  author={Jamieson, Kevin and Talwalkar, Ameet},
  booktitle={International Conference on Artificial Intelligence and Statistics},
  year={2016}
}

@article{li2018hyperband,
  title={Hyperband: A Novel Bandit-Based Approach to Hyperparameter Optimization},
  author={Li, Lisha and Jamieson, Kevin and DeSalvo, Giulia and Rostamizadeh, Afshin and Talwalkar, Ameet},
  journal={Journal of Machine Learning Research},
  volume={18},
  number={185},
  pages={1--52},
  year={2018}
}

@inproceedings{falkner2018bohb,
  title={{BOHB}: Robust and Efficient Hyperparameter Optimization at Scale},
  author={Falkner, Stefan and Klein, Aaron and Hutter, Frank},
  booktitle={International Conference on Machine Learning},
  year={2018}
}

@inproceedings{coreyes2021evolving,
  title={Evolving Reinforcement Learning Algorithms},
  author={Co-Reyes, John D and Miao, Yingjie and Peng, Daiyi and Real, Esteban and Le, Quoc V and Levine, Sergey and Lee, Honglak and Faust, Aleksandra},
  booktitle={International Conference on Learning Representations},
  year={2021}
}

@inproceedings{real2020automlzero,
  title={{AutoML-Zero}: Evolving Machine Learning Algorithms From Scratch},
  author={Real, Esteban and Liang, Chen and So, David R and Le, Quoc V},
  booktitle={International Conference on Machine Learning},
  year={2020}
}

@inproceedings{oh2020lpg,
  title={Discovering Reinforcement Learning Algorithms},
  author={Oh, Junhyuk and Hessel, Matteo and Czarnecki, Wojciech M and Xu, Zhongwen and van Hasselt, Hado and Singh, Satinder and Silver, David},
  booktitle={Advances in Neural Information Processing Systems},
  year={2020}
}

@incollection{lehman2022elm,
  title={Evolution Through Large Models},
  author={Lehman, Joel and Gordon, Jonathan and Jain, Shawn and Ndousse, Kamal and Yeh, Cathy and Stanley, Kenneth O},
  booktitle={Handbook of Evolutionary Machine Intelligence},
  year={2022},
  publisher={Springer}
}

@inproceedings{andrychowicz2021matters,
  title={What Matters in On-Policy Reinforcement Learning? A Large-Scale Empirical Study},
  author={Andrychowicz, Marcin and Raichuk, Anton and Sta{\'n}czyk, Piotr and Orsini, Manu and Girgin, Sertan and Marinier, Rapha{\"e}l and Hussenot, L{\'e}onard and Geist, Matthieu and Pietquin, Olivier and Michalski, Marcin and Gelly, Sylvain and Bachem, Olivier},
  booktitle={International Conference on Learning Representations},
  year={2021}
}

@inproceedings{burda2019exploration,
  title={Exploration by Random Network Distillation},
  author={Burda, Yuri and Edwards, Harrison and Storkey, Amos and Klimov, Oleg},
  booktitle={International Conference on Learning Representations},
  year={2019}
}

@misc{bou2023torchrl,
  title={{TorchRL}: A Data-Driven Deep Reinforcement Learning Library as an Out-of-the-Box Tool},
  author={Bou, Albert and Bettini, Matteo and Dittert, Sebastian and Kumar, Vikash and Sodhani, Shagun and Yang, Xiaomeng and De Fabritiis, Gianni and Moens, Vincent},
  year={2023},
  eprint={2306.00577},
  archivePrefix={arXiv},
  primaryClass={cs.LG}
}


\appendix

\section{Limitations}
\label{sec:limitations}

Three limitations bound what we claim. First, \emph{PPO and DQN results are suboptimal relative to what per-algorithm tuning would achieve}: budgets and hyperparameters were carried over from SAC without adjustment, so the absolute returns for these two families are not comparable to published results and represent a lower bound on what the same pipeline could reach with algorithm-specific calibration.

Second, \emph{the five-component decomposition is active only at L0}. Once a candidate advances to L1 or L2, it is evaluated as a monolithic program and the component interface plays no role. This confines Phase~1's per-component credit assignment, and any further structured exploration, to the 50K-step screening stage. A natural extension is to run L1 for a brief period and then re-enter L0 with the accumulated fitness signal---keeping structured five-component search active at longer horizons rather than switching permanently to monolithic evaluation.

Third, \emph{L0 screening at 50K steps can discard RL strategies with slow warm-up}. Programs that achieve strong asymptotic performance but require extended environment interaction to stabilise are ranked low at the short horizon and filtered out before the costlier evaluations that would reveal their advantage. The funnel's promotion thresholds are calibrated to the evaluation protocol, not to the learning dynamics of each candidate.

\section{Algorithm--Component Mapping}
\label{app:algo_map}

Table~\ref{tab:algo_component_map} shows how standard model-free algorithms populate the five slots of Section~\ref{sec:components}, supporting the claim that the decomposition is a template rather than a SAC-specific factoring.

\begin{table}[!htbp]
\centering
\footnotesize
\setlength{\tabcolsep}{4pt}
\caption{Standard model-free algorithms populate the same five slots; the decomposition is a template rather than a SAC-specific factoring. $^\ast$For value-based methods the policy slot is \emph{degenerate}, not absent: the policy is the greedy $\arg\max$ over $Q$, so the slot is fixed rather than searched (we therefore exclude it from the search for DQN). We evolve the three families marked in bold.}
\label{tab:algo_component_map}
\begin{tabular}{llllll}
\toprule
Algorithm & encode & value target & critic loss & policy loss & exploration \\
\midrule
\textbf{DQN}        & MLP & TD target      & MSE            & greedy$^\ast$      & $\epsilon$-greedy \\
Rainbow             & MLP & $n$-step       & distributional & greedy$^\ast$      & NoisyNet \\
DDPG / TD3          & MLP & twin TD target & MSE            & DPG                & OU / Gaussian \\
A2C / \textbf{PPO}  & MLP & GAE            & MSE            & clipped surrogate  & entropy \\
\textbf{SAC}        & MLP & soft Bellman   & MSE            & reparam.\ + entropy & entropy \\
\bottomrule
\end{tabular}
\end{table}

\section{Implications for RL Algorithm Design (extended)}
\label{app:discussion}

Our results suggest the standard SAC algorithm contains several recurring suboptimalities fixable across the majority of tasks tested: MSE is suboptimal for critic training (robust losses better on 16/17 tasks); fixed pessimistic Q-aggregation is overly conservative (annealing helps); raw observations benefit from logarithmic compression (symlog in 11/17); and constant-coefficient exploration bonuses are worse than progress-decayed ones. The approach holds across locomotion, manipulation (Meta-World 5/5), and driving (MetaDrive 3/3), consistent with cross-domain generality, and the PPO (5/5) and DQN (4/4) results obtained through the same five slots (Section~\ref{sec:cross_algo}) indicate the same holds across algorithm families.

\paragraph{Cross-domain generality.} In Meta-World, push and pick-place achieve order-of-magnitude gains from near-zero baselines, demonstrating that evolution can fundamentally restructure failing training procedures. In MetaDrive, all 3 driving scenarios improve, including 100-envs large-scale generalization (+19.0\%). The same evolution pipeline and evaluation protocol work across locomotion, robotic manipulation, and autonomous driving without domain-specific adjustments.

\paragraph{Cross-algorithm takeaway.} Across three algorithm families spanning off-policy and on-policy, continuous and discrete control, \sys improved the baseline on 26/26 tasks with audited, non-degenerate champions. Its two assumptions---that an RL algorithm decomposes into evolvable components, and that fitness can be estimated through a staged funnel---hold on all three, although the three are not validated to equal depth (Section~\ref{sec:limitations}).

\paragraph{Per-task failure modes of the single-level PPE variants.} L0-only costs $-$0.129 on dm-cheetah-run and $-$0.062 on mw-drawer-open; all-L2 costs $-$0.153 on md-100envs, the task with the most headroom.

\paragraph{Compositionality of the discovered innovations.} Huber loss appears in DQN (Mnih et al., 2015) and symlog in DreamerV3 (Hafner et al., 2023), so the individual ingredients are not new. Evolution combines them with elements absent from prior work (adaptive $\delta$ from running statistics, asymmetric weighting from Q-disagreement), and the specific schedules (3-stage Q-blending with sinusoidal interpolation) are novel.

\paragraph{Why the consensus innovations work.} Progressive Q-blending's annealing direction splits by domain: locomotion champions start optimistic and anneal toward $\min$, while manipulation and driving champions \emph{relax} pessimism as training stabilizes. Symlog encoding prevents gradient saturation in high-dimensional tasks.

\section{Implementation Details of PCE and PPE}
\label{app:impl}

\paragraph{The five component interfaces.} Each component is a Python function with a fixed calling convention, receiving the current step $t$ and total steps $T$ and sharing a persistent state dictionary; full Python signatures are included in the released code:
\begin{enumerate}[leftmargin=*,itemsep=1pt,topsep=2pt]
\item \textbf{\texttt{encode\_state}}: \emph{State representation} --- transforms raw observations into features (e.g., normalization, running statistics, dimensionality reduction).
\item \textbf{\texttt{compute\_value\_target}}: \emph{Value estimation} --- computes bootstrap targets for the critic. In SAC this is the soft Bellman target $\min(Q_1, Q_2) - \alpha\log\pi$; in PPO it is Generalized Advantage Estimation (GAE); in DQN the target-network Bellman backup.
\item \textbf{\texttt{compute\_critic\_loss}}: \emph{Critic learning} --- defines the loss for training value or Q-functions (e.g., MSE, Huber, value clipping).
\item \textbf{\texttt{compute\_policy\_loss}}: \emph{Policy optimization} --- computes the gradient signal for the policy. SAC uses reparameterized entropy-regularized objectives; PPO uses the clipped surrogate; DQN derives the policy implicitly via $\arg\max_a Q$.
\item \textbf{\texttt{compute\_exploration\_bonus}}: \emph{Exploration strategy} --- shapes the reward signal to encourage exploration (e.g., intrinsic curiosity, entropy bonuses, reward normalization).
\end{enumerate}

Algorithm~\ref{alg:pce} gives the full PCE procedure summarized in Section~\ref{sec:pce}, and Table~\ref{tab:ppe_levels} the three PPE fidelity levels of Section~\ref{sec:ppe}.

\begin{algorithm}[t]
\caption{Progressive Co-Evolution (PCE)}
\label{alg:pce}
\begin{algorithmic}[1]
\REQUIRE Base program $p_0$, components $\{m_1, \ldots, m_K\}$, candidate budget $B$ (240), batch size $b$ (10), L1 quota $Q$ (30)
\STATE \textbf{Phase 1}: bandit-guided single-component evolution
\WHILE{some arm is unfrozen \AND Phase-1 spend $<$ upper bound}
    \STATE Select arm $k^* = \argmax_{k \in \mathcal{K}_t} \text{UCB}(k)$ (Eq.~\ref{eq:ucb1}); generate batch of $b$ mutations of $m_{k^*}$; screen at L0
    \STATE Update bandit statistics $(\bar{r}_{k^*}, n_{k^*})$ with the batch's relative fitness gain
    \STATE Freeze arm $k^*$ if $\kappa$ consecutive pulls yielded relative gain $< 2\%$
\ENDWHILE
\STATE Greedily merge per-component champions into $p_{\text{merge}}$, keeping a component only if fitness does not regress
\STATE \textbf{Phase 2}: full co-evolution from $p_{\text{merge}}$ with the remaining budget $B - (\text{Phase-1 spend})$
\WHILE{budget remains}
    \STATE Run $b$ full-program mutations seeded with the current best program and the insight library
    \STATE Screen at L0; promote top-1 plus all candidates clearing gate $\gamma$ to L1 (Eq.~\ref{eq:ppe_promote}, within quota $Q$)
    \STATE Extract this round's distinguishing mechanisms into the insight library
\ENDWHILE
\STATE Evaluate the L1 pool's top-5 at L2
\RETURN Best program by audited L2 score
\end{algorithmic}
\end{algorithm}

\begin{table}[!htbp]
\centering
\small
\caption{PPE evaluation levels. Only programs passing lower levels are promoted.}
\label{tab:ppe_levels}
\begin{tabular}{lccccc}
\toprule
Level & Steps & Seeds & Eval Eps & Measured wall-clock & Purpose \\
\midrule
L0 & 50K & 1 & 5 & 8--11 min (9.3) & Rapid screening \\
L1 & 200K & 2 & 10 & 60--79 min (78) & Moderate validation \\
L2 & 1M & 3 & 20 & 4.1--9.6 h (5.0) & Final confirmation \\
\bottomrule
\end{tabular}
\end{table}

\paragraph{Evolution budget.} All mutations are generated by a large language model. Every run in this paper---main-line and ablation alike---draws from a budget of $B{=}240$ candidate programs. After syntactic validation and AST-hash deduplication, each run screens on average \textbf{230 candidates} at L0, for $\sim$87 GPU-hours of evaluation per run (Appendix~\ref{app:ppe}). The ablations of Section~\ref{sec:ablation} use the full procedure of Section~\ref{sec:pce} throughout, so all conditions compared there are internally consistent.

\paragraph{Matched-budget protocol for controlled comparisons.} The ablations of Section~\ref{sec:ablation} compare conditions under a matched mutation budget: every L0-screening condition receives 240 LLM-generated candidate programs, with the Phase-1/Phase-2 split determined adaptively as in Section~\ref{sec:pce}; conditions that remove a phase spend the same 240 candidates in the remaining phase. Conditions that evaluate every candidate at higher fidelity (L2-only, L1-only) cannot match candidate counts by construction, and are instead aligned by evaluation compute (same wall-clock GPU budget), which affords 13 (SAC), 43 (PPO) and 33 (DQN) L2-only candidates and 60 L1-only ones (realized $\approx$50 for L1-only due to evaluation time limits). The per-algorithm counts differ because the three trainers spend different numbers of environment steps per evaluation, so a fixed compute budget buys a different number of full-fidelity candidates in each; Appendix~\ref{app:ablation_full} records the alignment. All conditions run with isolated storage to prevent cross-condition contamination, and every reported champion passes the audit protocol of Section~\ref{sec:ppe} (loader verification of all five component functions plus AST comparison against the seed program).

\paragraph{Base SAC hyperparameters.} Our base SAC uses a 2-layer MLP (256 hidden), LayerNorm + ReLU, $\gamma{=}0.99$, $\tau{=}0.005$, batch size 256, learning rate $3{\times}10^{-4}$, automatic entropy tuning, 1M environment steps, and 5K random exploration steps. The external references (denisyarats/pytorch\_sac, CleanRL SAC) are run at identical network capacity and hyperparameters.

\paragraph{Evaluation protocol.} All results use L2: train with seeds $\{0, 42, 123\}$, then run deterministic evaluation episodes. For SAC and PPO this is 1M training steps and 20 evaluation episodes; for DQN, whose classic-control tasks have far shorter horizons, L2 is 500K steps and 30 evaluation episodes. Baselines and evolved programs of a given algorithm always use that algorithm's L2 setting, so every comparison in this paper is within-protocol. We report mean return $\pm$ std across seeds. To guard against silent failures in the evolution pipeline, every reported champion program was additionally \emph{audited}: we re-load the program with the component loader, verify that all five component functions are extractable, and confirm the reported fitness reproduces at L2. Only audited programs are reported.

\paragraph{Provenance of the reported numbers.} In Table~\ref{tab:main_results}, the DMControl and HalfCheetah entries of the SAC blocks are independent 3-seed re-evaluations, and the Hopper/Walker2d/Humanoid entries are Aug 2026 re-evaluations. For the PPO and DQN blocks of the same table, the evolution pipeline persisted only a scalar L2 fitness for the champions, so the evolved entries are \emph{independent} re-evaluations of the champion programs under exactly the protocol their baselines use. Because each champion was selected as the maximum over $N$ candidates, its recorded selection score is upward-biased; the re-evaluations therefore land slightly below it on 7 of 9 tasks (by $0.001$--$0.071$ fitness) and above it on the other two. We report the re-evaluation rather than the selection score throughout.

\paragraph{Why MountainCar-v0 uses 10 seeds.} Its outcome is bimodal---a run either solves the task at $\approx-130$ or fails at exactly $-200$---so a 3-seed mean is unstable: an earlier 3-seed draw caught two failures and reported $-178.6$ for the same champion that scores $-144.9$ over 10 seeds. Both sides of that comparison use 10 seeds. Over those 10, the baseline returns exactly $-200.0$ on every seed (std $=0$) and the evolved program reaches the goal on 8, which is the 0/10 $\to$ 8/10 success-rate shift quoted in Section~\ref{sec:cross_algo}.

\section{Full Per-Task Results}
\label{app:full_results}

\paragraph{SOTA-init effect sizes relative to seed noise.} Section~\ref{sec:sota_init} reports that six of the nine SOTA-init gains move by less than a single combined seed spread (root-sum-square of the two cells' standard deviations). The per-task multiples are: Walker2d $0.04\times$, HalfCheetah $0.24\times$, Hopper $0.58\times$, quadruped-walk $0.71\times$, Humanoid $0.94\times$, cheetah-run $0.99\times$, Ant $1.34\times$, walker-walk $1.38\times$, dm-walker-run $2.22\times$. Each is computed from the baseline and evolved cells of the SOTA-init block below.

The full per-task table appears as Table~\ref{tab:main_results} in the main body; it is not duplicated here. Its caption there carries the protocol, the marker definitions ($\dagger$, $^\S$, $^\P$) and the per-cell provenance.

\paragraph{Per-task SOTA-init gaps and the head-to-head outcome.} The 9-task mean gap between the two seed implementations falls from $23.4\%$ to $13.1\%$ in return (Section~\ref{sec:sota_init}). Measured against the SOTA seed's own return, the per-task gap goes: walker-walk $29.8\%\to0.2\%$, walker-run $35.2\%\to24.2\%$, cheetah-run $57.6\%\to24.4\%$, quadruped-walk $0.7\%\to0.1\%$, HalfCheetah $53.7\%\to18.2\%$, Hopper $4.8\%\to1.8\%$, Walker2d $16.9\%\to15.8\%$, Humanoid $3.2\%\to6.4\%$, Ant $8.4\%\to27.0\%$. These are mean absolute deviations with the SOTA initiated seed as the denominator, and the mean of the first entry in each pair is the $23.4\%$ quoted above; the second entries mean to $13.1\%$. The gap narrows on 7 of 9. It widens only on Humanoid and Ant, the two cells whose \emph{starting} gap is among the smallest ($3.2\%$ and $8.4\%$), so there is little room to close and the sign is within the run-to-run spread of those cells.

Table~\ref{tab:sota_gaps} puts the two quantities side by side: how far apart the seeds were, and what each batch's own search subsequently did. Columns \emph{own gain} and \emph{SOTA gain} are each batch's improvement relative to \emph{its own} seed, so they are the mechanism behind the \emph{outcome} column rather than a second measurement of the gap.

\begin{table}[t]
\centering
\caption{The nine locomotion tasks both SAC batches were evolved on with 3 seeds. \emph{Seed gap}: the SOTA-init baseline relative to the own-init baseline. \emph{own gain} / \emph{SOTA gain}: each batch's L2 champion relative to its own seed, i.e.\ what that batch's search contributed. \emph{Final gap} and \emph{outcome}: the two champions against each other, with the SOTA-init champion as denominator throughout. All figures are on raw returns, since fitness is normalised per task and the batches share a normalisation only within a task; a positive value always means the SOTA-init side is ahead. The final-gap column disaggregates the $23.4\%\to13.1\%$ mean narrowing; the mean of its absolute values is $13.1\%$. Note that ``SOTA-init'' names the batch, not a claim that its seed is stronger on every task---it starts behind on quadruped-walk and Hopper.}
\label{tab:sota_gaps}
\label{app:sota_gaps}
\setlength{\tabcolsep}{4pt}
\renewcommand{\arraystretch}{1.15}
\begin{tabular*}{\textwidth}{@{}@{\extracolsep{\fill}}lrrrrl@{}}
\toprule
Task & Seed gap & own gain & SOTA gain & Final gap & Outcome \\
\midrule
dm-cheetah-run      & $+57.6\%$  & $+114.0\%$ & $+20.0\%$  & $+24.4\%$  & SOTA-init \\
dm-walker-walk      & $+29.8\%$  & $+42.9\%$  & $+0.5\%$   & $+0.2\%$   & tie \\
dm-walker-run       & $+35.2\%$  & $+20.1\%$  & $+2.6\%$   & $+24.2\%$  & SOTA-init \\
dm-quadruped-walk   & $-0.7\%$   & $+0.9\%$   & $+1.7\%$   & $+0.1\%$   & tie \\
gym-HalfCheetah-v5  & $+53.7\%$  & $+77.7\%$  & $+0.6\%$   & $+18.2\%$  & SOTA-init \\
gym-Hopper-v5       & $-4.8\%$   & $+12.4\%$  & $+20.0\%$  & $+1.8\%$   & SOTA-init \\
gym-Walker2d-v5     & $+16.9\%$  & $+39.8\%$  & $+0.4\%$   & $-15.8\%$  & own-init \\
gym-Humanoid-v5     & $+3.2\%$   & $+5.6\%$   & $+9.3\%$   & $+6.4\%$   & SOTA-init \\
gym-Ant-v5          & $+8.4\%$   & $+1.4\%$   & $+27.3\%$  & $+27.0\%$  & SOTA-init \\
\midrule
\multicolumn{6}{@{}l@{}}{\textit{Mean $|\text{seed gap}|$ $23.4\%$, mean $|\text{final gap}|$ $13.1\%$; mean own gain $+35.0\%$, mean SOTA gain $+9.2\%$.}}\\
\bottomrule
\end{tabular*}
\end{table}

Three readings follow. First, narrowing and outcome are different questions, and here they come apart: the gap narrows on 7 of 9 tasks, but the batch that starts ahead finishes ahead on only 4 of those 7. Quadruped-walk and Hopper both cross over---own-init starts marginally ahead and finishes marginally behind---though at $0.7\%\to0.1\%$ and $4.8\%\to1.8\%$ both crossings stay within a few points of zero. What separates the cases is the size of the starting advantage: all four preserved orderings sit on seed gaps of $29.8\%$ or more, and every task where the seed gap was under $17\%$ changed hands. A large seed advantage is durable; a small one is not.

Second, the only ordering change of any magnitude is Walker2d, and the \emph{gains} explain it rather than the seeds. There the own-init search contributes $+39.8\%$ against the SOTA-init search's $+0.4\%$, so the batch starting $16.9\%$ behind overtakes and finishes $15.8\%$ ahead. Ant is the mirror case, with the SOTA-init search contributing $+27.3\%$ against $+1.4\%$---the same cell identified in Section~\ref{sec:sota_init} as the one where own-init search failed, which is why its gap widens rather than narrows. In both, the outcome follows the per-task gain, not the per-task seed.

Third, the gains are far more uneven than the seed gaps: they span $+0.4\%$ to $+114.0\%$, while the seed gaps matter only up to about $58\%$ and, below $17\%$, not at all. Mean own-init gain is $+35.0\%$ against $+9.2\%$ for SOTA-init---expected, since a stronger seed leaves less headroom---but the per-task spread of that second column (from $+0.4\%$ to $+27.3\%$) is what actually decides which batch finishes ahead on the close tasks.

\section{Per-Family Qualifications}
\label{app:family_notes}

\paragraph{Cross-family comparisons are raw-return comparisons only.} Fitness is normalised by task-specific bounds, and the bounds are shared across families, but the \emph{scale} of the return differs by orders of magnitude between DQN's classic-control tasks (CartPole ceiling 500) and the locomotion tasks (HalfCheetah in the $10^4$ range). A per-family median gain is therefore a statement about each family's own task suite, not a ranking of the families' algorithms; the three medians ($+39.8\%$, $+83.6\%$, $+32.1\%$) are not comparable to each other in magnitude, and none is comparable to published numbers, which use different evaluation protocols. The one cross-family comparison we do make---evolved PPO against our evolved SAC on gym-Hopper-v5---is made on raw return under one protocol (Section~\ref{sec:cross_algo}), which is why it is stated that way and not as a fitness difference.

\paragraph{The marker definitions.} Three markers appear in Table~\ref{tab:main_results}. $^\dagger$ marks an own-init SAC cell whose evolved return exceeds the external open-source reference for that task (denisyarats/pytorch\_sac on DMControl, CleanRL SAC on Gymnasium) as we re-evaluated it at identical capacity and hyperparameters. $^\S$ (on the group headers) flags that the evolved entries of that block are \emph{independent} re-evaluations of the champion programs rather than their selection scores, which are upward-biased by max-over-$N$ selection; the same applies to the two SAC blocks, where the re-evaluation provenance is recorded per cell in Appendix~\ref{app:impl}. $^\P$ marks MountainCar-v0, evaluated with 10 seeds on both sides rather than 3; the reason and the full draw are in Appendix~\ref{app:perseed}.

\paragraph{The $\dagger$ comparisons.} Two qualifications attach to a $\dagger$. First, the reference numbers are our re-evaluations, not the values published by those projects, so a $\dagger$ means we beat our own reproduction of the reference under our protocol, not that we beat a leaderboard entry. Second, all five $\dagger$ cells sit in the own-init block, where the comparison crosses a change of seed program as well as a change of scale---five of the nine locomotion tasks are cleared from our own weaker seed (walker-walk, quadruped-walk, Hopper, Walker2d, Humanoid), and the SOTA-init block re-runs the same nine to isolate the scale effect, where the margins are much smaller (median $+2.6\%$).

\paragraph{The MountainCar normalisation artifact.} MountainCar-v0's baseline scores a fitness of $0.908$ while returning exactly $-200.0$ on every seed, i.e.\ the worst attainable return for that environment. The task's normalisation bounds place $-200$ near the top of the fitness range, so the worst possible policy receives a high fitness. This is why the DQN ablation reports all conditions within $0.908$--$0.948$ on that task (Section~\ref{sec:ablation}) and why we read no ordering from that column: it compresses the entire failure region into a narrow high-fitness band, and the task cannot separate conditions there. The relevant MountainCar result is the success-rate shift under 10 seeds ($0/10\to8/10$), not any fitness gap, and we report it that way.

\section{Benchmark Task Details}
\label{app:tasks}

\begin{table}[!htbp]
\centering
\caption{SAC benchmark tasks with observation/action dimensions and normalization bounds. All 17 are evolved in the own-init batch; the 9 locomotion tasks (all DMControl and all Gymnasium entries) additionally have an external SOTA reference and were also run in the SOTA-init batch, all 9 of which are reported in Table~\ref{tab:main_results}. The cross-algorithm experiments additionally use gym-Hopper/HalfCheetah/Ant/Walker2d/Humanoid-v5 for PPO and CartPole-v1, MountainCar-v0, Acrobot-v1, LunarLander-v3 for DQN.}
\label{tab:tasks}
\begin{tabular}{llccc}
\toprule
Task & Source & Obs Dim & Act Dim & Norm Bound \\
\midrule
dm-cheetah-run & DMControl & 17 & 6 & 1000 \\
dm-walker-walk & DMControl & 24 & 6 & 1000 \\
dm-walker-run & DMControl & 24 & 6 & 1000 \\
dm-quadruped-walk & DMControl & 78 & 12 & 1000 \\
\midrule
gym-HalfCheetah-v5 & Gymnasium & 17 & 6 & 15000 \\
gym-Hopper-v5 & Gymnasium & 11 & 3 & 4000 \\
gym-Walker2d-v5 & Gymnasium & 17 & 6 & 6000 \\
gym-Ant-v5 & Gymnasium & 105 & 8 & 8000 \\
gym-Humanoid-v5 & Gymnasium & 348 & 17 & 7000 \\
\midrule
mw-push-v3 & Meta-World & 39 & 4 & 4500 \\
mw-drawer-open-v3 & Meta-World & 39 & 4 & 4500 \\
mw-door-open-v3 & Meta-World & 39 & 4 & 4500 \\
mw-peg-insert-side-v3 & Meta-World & 39 & 4 & 4500 \\
mw-pick-place-v3 & Meta-World & 39 & 4 & 4500 \\
\midrule
md-validation & MetaDrive & 259 & 2 & 350 \\
md-10env & MetaDrive & 259 & 2 & 350 \\
md-100envs & MetaDrive & 259 & 2 & 350 \\
\bottomrule
\end{tabular}
\end{table}

\section{Cross-Task Generalization}
\label{app:generalization}

A critical question is whether evolved programs are overfit to their training task or discover generalizable improvements. Table~\ref{tab:cross_task} gives the full transfer matrix in numeric form: 4 champion programs (evolved on dm-walker-walk, dm-quadruped-walk, gym-Walker2d-v5, and gym-Humanoid-v5) re-evaluated under the same 3-seed L2 protocol as Table~\ref{tab:main_results} on 9 locomotion targets spanning DMControl and Gymnasium, alongside each target's fixed baseline fitness so that every $\Delta$ can be read as a fraction of its baseline. Figure~\ref{fig:cross_task} in the main body renders the same 36 cells as a heatmap.

\begin{table}[!htbp]
\centering
\small
\setlength{\tabcolsep}{4.5pt}
\caption{Cross-task L2 generalization, full $9\times4$ matrix: fitness improvement ($\Delta$) over the fixed L2 baseline when a champion evolved on one task is re-evaluated on each target. Source task marked with *; the summary rows exclude source cells. Bold = best source program for that target. \emph{Provenance.} The matrix was evaluated as a single batch using the champions available at that time. The gym-Walker2d-v5 own-init champion was subsequently improved by resumed evolution (fitness $0.760 \rightarrow 0.905$), so the entire walker2d column describes the \emph{earlier} program. We did not substitute the newer champion into individual cells: only its own-task value was re-measured, so patching the diagonal alone would leave a column mixing two different programs and therefore not internally comparable. Two consequences follow. The walker2d diagonal here ($0.647+0.113 = 0.760$) is below the gym-Walker2d-v5 own-init row of Table~\ref{tab:main_results} ($0.905$); and the weak transfer of that column is a property of the earlier program, \emph{not} an established property of the champion reported in Table~\ref{tab:main_results}.}
\label{tab:cross_task}
\begin{tabular}{lccccc}
\toprule
 & \multicolumn{4}{c}{Source program ($\Delta$ fitness)} & \\
\cmidrule(lr){2-5}
Task & walker-walk* & quadruped* & walker2d* & Humanoid* & Baseline Fitness \\
\midrule
dm-cheetah-run & +0.265 & +0.269 & $-$0.129 & \textbf{+0.313} & 0.302 \\
dm-walker-walk & \textbf{+0.292}* & +0.290 & +0.243 & +0.221 & 0.680 \\
dm-walker-run & +0.091 & +0.025 & $-$0.319 & \textbf{+0.142} & 0.488 \\
dm-quadruped-walk & \textbf{+0.019} & +0.006* & $-$0.669 & $-$0.082 & 0.932 \\
\midrule
gym-HalfCheetah-v5 & +0.058 & \textbf{+0.154} & $-$0.148 & +0.141 & 0.340 \\
gym-Hopper-v5 & $-$0.207 & $-$0.076 & $-$0.001 & \textbf{+0.201} & 0.619 \\
gym-Walker2d-v5 & +0.028 & +0.003 & +0.113* & \textbf{+0.163} & 0.647 \\
gym-Ant-v5 & \textbf{+0.239} & $-$0.056 & +0.188 & +0.155 & 0.467 \\
gym-Humanoid-v5 & \textbf{+0.279} & +0.052 & $-$0.234 & +0.056* & 0.714 \\
\midrule
Cross-task wins & 7/8 & 6/8 & 2/8 & \textbf{7/8} & --- \\
Mean $\Delta$ (excl. source) & +0.097 & +0.083 & $-$0.134 & \textbf{+0.157} & --- \\
\bottomrule
\end{tabular}
\end{table}

Counting only non-source cells, the \textbf{Humanoid program} generalizes best: 7/8 targets improved at mean $+0.157$ fitness, and it is the best source program on 4 of the 9 rows. The \textbf{walker-walk program} ties it on win count (7/8) at a smaller mean ($+0.097$) despite being evolved on the simplest task in the suite, and is likewise best on 4 rows; its single failure is gym-Hopper-v5 ($-0.207$). The \textbf{quadruped program} improves 6/8 ($+0.083$); its modifications (symlog encoding, cosine-annealed Q blending, Log-Cosh critic loss) are broadly beneficial but lose on gym-Hopper-v5 and gym-Ant-v5. The \textbf{walker2d program overfits}: 2/8 wins at mean $-0.134$, including $-0.669$ on dm-quadruped-walk, relying on step-dependent strategies with hard-coded thresholds. Two caveats bound how far this ordering can be pushed. Each column is a single program, so differences between columns conflate the source task with the particular champion that task happened to produce; and the walker2d column describes a program that later evolution superseded (see caption), so its poor transfer is not established for the champion of Table~\ref{tab:main_results}. The principles behind the difference are distilled below.

\paragraph{What makes programs generalizable?} Comparing a well-generalizing program (quadruped, 6/8 wins) with the worst (walker2d, 2/8 wins), we identify three principles for transferable RL modifications. \textbf{(1) Relative scheduling}: use progress ratio $t/T$ rather than absolute step counts for annealing---the quadruped program's cosine-annealed schedules adapt to any training length. \textbf{(2) Stateless encoding}: pointwise transformations (arcsinh, symlog) transfer perfectly; running statistics do not. \textbf{(3) Adaptive thresholds}: parameters derived from training dynamics (EMA of TD errors, as in the walker-walk program) auto-calibrate to each task, while hard-coded constants may be inappropriate for different reward scales.

\section{Discovered Algorithmic Innovations: Detailed Analysis}
\label{app:innovations}

\paragraph{Huber loss addresses bootstrapping noise.} SAC's critic is trained on bootstrapped targets that are inherently noisy early in training. MSE amplifies large errors quadratically, destabilizing learning. Huber loss caps the gradient at $\delta$, providing natural gradient clipping without explicit hyperparameter tuning.

\paragraph{Q-annealing balances exploration-exploitation.} The pessimism of $\min(Q_1, Q_2)$ underestimates state values, which can suppress exploration. Early optimism (toward mean) encourages visiting novel states; progressive conservatism later prevents overestimation from corrupting the policy.

\paragraph{Symlog prevents saturation.} In tasks like MetaDrive (259D LiDAR observations) or HalfCheetah (high-velocity states), raw observations can reach extreme values that saturate ReLU networks. Logarithmic compression keeps inputs in a network-friendly range.

\paragraph{Robust critic loss (16/17 tasks; EMA-adaptive scale in 9).} The most universal discovery replaces MSE with a robust loss (Huber, smooth-L1, Log-Cosh, or Charbonnier); 9 champions further adapt the transition scale from running error statistics---8 as a Huber $\delta$, and md-10env as a Log-Cosh $\kappa$ clamped to $[1,5]$:

\begin{lstlisting}[style=python, caption={Evolved adaptive Huber loss (dm-walker-walk, simplified)}]
# Track TD-error distribution with EMA
td_errors = (q_pred - q_target).abs()
self.ema_delta = 0.99 * self.ema_delta + 0.01 * td_errors.quantile(0.9)

# Asymmetric weighting based on Q-disagreement
q_disagreement = (q1 - q2).abs().mean()
tau = 0.55 + 0.2 * sigmoid(q_disagreement - self.ema_disagreement)

# Asymmetric Huber: overestimation penalized more heavily
residual = q_pred - q_target
loss = torch.where(
    residual.abs() < self.ema_delta,
    0.5 * residual**2 / self.ema_delta,
    residual.abs() - 0.5 * self.ema_delta
)
loss = torch.where(residual > 0, tau * loss, (1-tau) * loss)
\end{lstlisting}

\paragraph{Progressive Q-blending (12/17 tasks).} Standard SAC uses $\min(Q_1, Q_2)$ (pessimistic). Evolution discovers progress-annealed blending of $\min$ and mean. A representative locomotion schedule (dm-walker-walk): early training ($t/T < 0.3$) uses $\text{mean}(Q_1, Q_2)$ for optimistic exploration, shifting via sinusoidal annealing to pure $\min$ for conservative exploitation. Manipulation and driving champions typically anneal in the opposite direction, relaxing pessimism as training stabilizes.

\paragraph{Symlog encoding (11/17 tasks).} Evolution discovers $\text{encode}(x) = \text{sign}(x) \cdot \log(1 + |x|)$ or equivalently $\text{arcsinh}(x/\sigma) \cdot \sigma$, preserving sign while compressing dynamic range. This was independently proposed in DreamerV3 (Hafner et al., 2023).

\section{PPE Efficiency}
\label{app:ppe}

Table~\ref{tab:ppe_stats} quantifies PPE's filtering across the 6 locomotion tasks. Each run evaluates 230 L0 candidates; evaluating all of them directly at L2 ($\sim$5.8 GPU-hours each) would require $\sim$1330 GPU-hours per run. With PPE, L0 screening costs $\sim$37 GPU-hours, the surviving $\sim$18 candidates cost $\sim$21 GPU-hours at L1, and the final 5 candidates cost $\sim$29 GPU-hours at L2---roughly 87 GPU-hours in total, a \textbf{15$\times$ reduction}. The L0$\to$L1 promotion rate of $\approx$7.6\% means PPE eliminates 92\% of candidates at minimal cost.

\begin{table}[!htbp]
\centering
\small
\caption{PPE filtering efficiency across 6 locomotion tasks. Statistics are per run, over the full $B{=}240$ candidate budget.}
\label{tab:ppe_stats}
\begin{tabular*}{0.85\textwidth}{@{\extracolsep{\fill}}lccc}
\toprule
Metric & Mean & Min & Max \\
\midrule
L0 evaluations per run & 230 & 230 & 230 \\
L0$\to$L1 promotion rate & 7.6\% & 4.8\% & 12.2\% \\
Code deduplication rate & 18\% & 2.5\% & 30\% \\
\multicolumn{4}{@{}l@{}}{\footnotesize\emph{Fraction of generated candidate programs discarded as AST-duplicates before screening;}} \\
\multicolumn{4}{@{}l@{}}{\footnotesize\emph{the denominator is the $B{=}240$ budget, so $\approx$230 survive to L0.}} \\
L1 evaluations per run & 17.5 & 11 & 28 \\
L2 evaluations per run & 5 & 5 & 5 \\
GPU-hours per run (L0 / L1 / L2) & 36.8 / 21.1 / 28.9 & --- & --- \\
\bottomrule
\end{tabular*}
\end{table}

\begin{figure}[!htbp]
\centering
\includegraphics[width=0.86\textwidth]{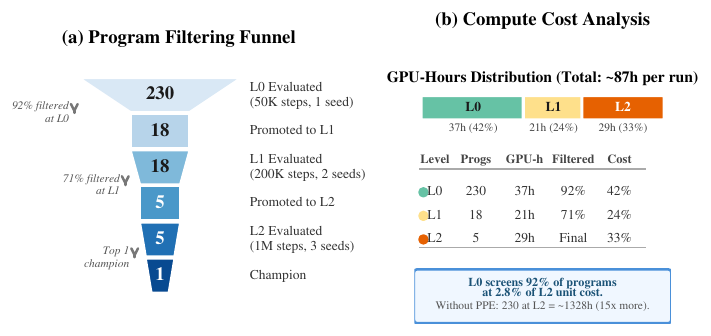}
\caption{PPE filtering efficiency (per-run averages). \textbf{(a)} Program filtering funnel: from 230 L0-evaluated candidates to 5 L2-validated finalists. L0 filters 92\% of programs at minimal cost. \textbf{(b)} GPU compute distribution: total $\sim$87 GPU-hours per run versus $\sim$1330 for evaluating all candidates at L2, a $\sim$15$\times$ reduction.}
\label{fig:ppe_funnel}
\end{figure}

\paragraph{Cross-level rank agreement.} The filtering claim of Section~\ref{sec:related} rests on the assumption that low-fidelity returns rank candidates in roughly the order their L2 returns would. We test that directly rather than assuming it. Pooled over the ablation runs (A0, A1 and w/o Phase~1; $n{=}466$ candidate triplets) the Spearman correlation with L2 fitness is $+0.80$ for L1 ($p\approx2{\times}10^{-105}$, block-bootstrap 95\% CI $[+0.72,+0.86]$, resampling whole runs) against $+0.73$ for L0, and the L1 figure is stable across all three conditions ($+0.77$/$+0.79$/$+0.83$). L1 is therefore a sound proxy for L2, which is what justifies spending L1 budget on the $\sim$18 survivors rather than promoting straight from L0. The L0 figure is weaker evidence: much of its pooled $+0.73$ comes from cross-task scale differences rather than from within-run ordering, and the correct within-run reading is confined to single-layer conditions. We report the pooled correlations as a check on the funnel's ordering, not as a claim that L0 ranking is reliable.

\begin{center}
\small
\begin{tabular}{lccc}
\toprule
Correlation with L2 fitness & $\rho$ & $n$ & 95\% CI \\
\midrule
L1 $\to$ L2 (A0 + A1 + w/o Phase~1) & $+0.801$ & 466 & $[+0.72, +0.86]$ \\
L0 $\to$ L2 (same pool)              & $+0.727$ & 466 & --- \\
L1 $\to$ L2, A0 only                 & $+0.774$ & 160 & --- \\
L1 $\to$ L2, A1 only                 & $+0.790$ & 160 & --- \\
L1 $\to$ L2, w/o Phase~1 only        & $+0.830$ & 146 & --- \\
\bottomrule
\end{tabular}
\end{center}

\section{Full Per-Seed Results}
\label{app:perseed}

Table~\ref{tab:baseline_perseed} gives the per-seed L2 returns behind every mean $\pm$ std in Table~\ref{tab:main_results}, for all 17 SAC own-init tasks plus the 5 PPO and 4 DQN tasks. Seeds are the training seeds $\{0,42,123\}$ under each algorithm's own L2 protocol (SAC/PPO 1M steps and 20 episodes; DQN 500K steps and 30 episodes); the $\pm$ in Table~\ref{tab:main_results} is the population standard deviation of these three numbers. The spread is highly task-dependent, and reading the individual seeds changes how several headline rows should be interpreted. Within SAC, the two Meta-World tasks with order-of-magnitude gains are also the two least reliable: mw-pick-place-v3 solves the task on one seed out of three (2887.3 vs.\ 2.9 and 3.9) and mw-peg-insert-side-v3 on two out of three; those rows should be read as ``evolution finds a solution that sometimes trains'' rather than as a stable improvement. In DQN, the evolved CartPole program hits the environment's 500-return ceiling on two of three seeds, so its $\pm 69.0$ is one-sided censoring rather than symmetric noise, and the LunarLander baseline spans $-31.6$ to $180.0$ (it crashes on the first seed) while all three evolved seeds land---which is the substance of that row's improvement, more than the change in mean. PPO's per-seed spreads are otherwise unremarkable, with gym-Hopper-v5 the tightest cell in the whole table (3575.1 / 3444.4 / 3556.9).

\begin{table}[!htbp]
\centering
\small
\setlength{\tabcolsep}{4pt}
\caption{Per-seed L2 returns for all 17 SAC own-init tasks and all 9 PPO/DQN tasks (training seeds $\{0,42,123\}$; SAC/PPO 1M steps and 20 evaluation episodes, DQN 500K steps and 30 episodes). These are the values summarized in Table~\ref{tab:main_results}. $\P$~MountainCar-v0 is the one exception: because its outcome is bimodal (solve at $\approx-130$ or fail at $-200$), Table~\ref{tab:main_results} reports a 10-seed re-evaluation rather than a 3-seed one. The three seeds shown here are the $\{0,42,123\}$ subset of that draw; the full 10 evolved returns are $-120.7$, $-141.2$, $-200.0$, $-153.9$, $-124.8$, $-200.0$, $-128.8$, $-124.4$, $-134.7$, $-120.6$ (mean $-144.9$), and the baseline returns exactly $-200.0$ on all 10. The 3-seed subset therefore does \emph{not} average to the reported mean.}
\label{tab:baseline_perseed}
\begin{tabular*}{\textwidth}{@{\extracolsep{\fill}}lrrrcrrr}
\toprule
& \multicolumn{3}{c}{Baseline} & & \multicolumn{3}{c}{Evolved} \\
\cmidrule{2-4}\cmidrule{6-8}
Task & Seed 0 & Seed 42 & Seed 123 & & Seed 0 & Seed 42 & Seed 123 \\
\midrule
\multicolumn{8}{l}{\textit{SAC --- DMControl}} \\
dm-cheetah-run & 344.0 & 309.4 & 253.5 & & 612.4 & 741.1 & 587.3 \\
dm-walker-walk & 626.0 & 687.4 & 725.5 & & 974.3 & 972.4 & 966.5 \\
dm-walker-run & 553.1 & 415.6 & 494.4 & & 583.7 & 595.2 & 578.1 \\
dm-quadruped-walk & 936.2 & 914.5 & 944.0 & & 938.2 & 935.8 & 945.4 \\
\midrule
\multicolumn{8}{l}{\textit{SAC --- Gymnasium}} \\
gym-HalfCheetah-v5 & 5255.9 & 4798.0 & 5226.8 & & 11164.8 & 7468.8 & 8527.6 \\
gym-Hopper-v5 & 2441.5 & 1740.7 & 3242.8 & & 3256.9 & 1571.4 & 3514.0 \\
gym-Walker2d-v5 & 4044.3 & 4030.5 & 3574.6 & & 5797.4 & 4743.7 & 5747.3 \\
gym-Humanoid-v5 & 5261.6 & 4974.8 & 4760.0 & & 5484.6 & 5116.7 & 5242.3 \\
gym-Ant-v5 & 4458.2 & 3439.8 & 3300.5 & & 3129.7 & 3283.1 & 4944.8 \\
\midrule
\multicolumn{8}{l}{\textit{SAC --- Meta-World}} \\
mw-push-v3 & 5.6 & 4.7 & 4.8 & & 240.2 & 4459.3 & 776.0 \\
mw-drawer-open-v3 & 1493.1 & 3988.8 & 4049.1 & & 4562.4 & 4607.1 & 4694.5 \\
mw-door-open-v3 & 3621.1 & 3445.1 & 1651.8 & & 380.6 & 4481.8 & 4437.4 \\
mw-peg-insert-side-v3 & 25.7 & 5.3 & 3424.7 & & 4607.0 & 2.7 & 4617.0 \\
mw-pick-place-v3 & 6.8 & 9.8 & 3.9 & & 2.9 & 2887.3 & 3.9 \\
\midrule
\multicolumn{8}{l}{\textit{SAC --- MetaDrive}} \\
md-validation & 266.1 & 112.5 & 65.0 & & 459.9 & 395.8 & 114.0 \\
md-10env & 304.2 & 164.0 & 220.7 & & 223.2 & 205.3 & 269.5 \\
md-100envs & 311.7 & 164.4 & 235.1 & & 254.4 & 282.9 & 309.2 \\
\midrule
\multicolumn{8}{l}{\textit{PPO --- Gymnasium}} \\
gym-Hopper-v5 & 2128.4 & 1510.0 & 2121.6 & & 3575.1 & 3444.4 & 3556.9 \\
gym-HalfCheetah-v5 & 1682.9 & 1712.5 & 1777.6 & & 3837.5 & 4175.1 & 4688.3 \\
gym-Ant-v5 & 795.4 & 858.1 & 674.4 & & 2411.5 & 2541.7 & 3413.9 \\
gym-Walker2d-v5 & 4229.1 & 3138.4 & 4702.6 & & 4357.8 & 3899.9 & 5062.8 \\
gym-Humanoid-v5 & 561.7 & 573.9 & 684.2 & & 938.8 & 744.4 & 744.0 \\
\midrule
\multicolumn{8}{l}{\textit{DQN --- classic control}} \\
CartPole-v1 & 311.9 & 354.8 & 324.0 & & 500.0 & 500.0 & 353.6 \\
MountainCar-v0$^\P$ & $-$200.0 & $-$200.0 & $-$200.0 & & $-$120.7 & $-$141.2 & $-$200.0 \\
LunarLander-v3 & $-$31.6 & 52.2 & 180.0 & & 150.6 & 165.4 & 207.8 \\
Acrobot-v1 & $-$92.8 & $-$84.1 & $-$96.8 & & $-$80.1 & $-$84.9 & $-$95.9 \\
\bottomrule
\end{tabular*}
\end{table}

\section{Evaluation Wall-Clock Caps and Their Effect on the Ablation}
\label{app:eval_caps}

Every evaluation runs under a per-level wall-clock cap, and every run under a total cap. Because the ablation conditions differ in how expensive a single evaluation is, these caps do not bind equally, and Section~\ref{sec:limitations} summarises the consequence. The per-condition figures are as follows.

\textbf{(1)~L0 cap (1200\,s).} It truncates 40--65\% of PPO L0 evaluations. All L0-screening conditions are affected equally, so there is no condition-level selection bias, but the L0 signal is noisier than it would otherwise be.

\textbf{(2)~Headroom differs by condition.} The L2-only cap (43200\,s) is $7.3\times$ its median evaluation cost and is therefore structurally immune to truncation, whereas L0-only ($1.95\times$ headroom) and L1-only ($1.76\times$) each lose roughly half their samples under slow-tail conditions. Part of L2-only's apparent advantage in Table~\ref{tab:ablation_full} therefore has a structural rather than methodological origin.

\textbf{(3)~SAC L1-only is right-censored.} It is the only condition for which the 4800\,s L1 cap binds, so its Table~\ref{tab:ablation_full} entry is a right-censored estimate of what an uncapped L1-only run would reach.

\textbf{(4)~Run-level cap (96\,h).} This truncated 4 of the 6 PPO L2-only runs across both PPO pairs (35/42/34/36 of 43 planned L2 evaluations); all DQN L2-only runs reached their full 33. Truncation removes candidates from L2-only and thus works \emph{against} it, so the reported PPO L2-only margin over the full method is a conservative lower bound.

\section{Full Component-Ablation Matrix}
\label{app:ablation_full}

Table~\ref{tab:ablation_main} summarises the component ablation; Table~\ref{tab:ablation_full} gives every cell. For completeness we also report the exact two-sided Wilcoxon signed-rank test of each ablation against the full method over the 8 paired differences, in the column order of Table~\ref{tab:ablation_full}: $p = 0.383/0.250/0.195/0.383/0.312$. The smallest value attainable at $n{=}8$ is 0.008, so this design has little power against effects of the observed size; we report the paired $\Delta \pm$ std in the main text for that reason. Mean ranks over the six conditions within each pair (average-rank convention for ties) are $3.06/3.50/3.69/4.38/2.88/3.50$ over all 8 pairs and $2.50/3.25/3.75/4.33/3.50/3.67$ over the 6 non-degenerate pairs: the full method is second to L2-only on the first grouping and first on the second. Two cells are degenerate and marked $^{*}$: PPO on gym-HalfCheetah-v5, whose low values reflect PPO's own weakness on that task under the shared normalisation rather than any ablation effect, and MountainCar-v0, which is saturated (Appendix~\ref{app:ablation_dqn}). Neither enters the SAC mean, nor the 6-pair rows of Table~\ref{tab:ablation_main}. Because the matrix is complete, the \emph{differences} along the 8-pair mean row are identical to the mean paired $\Delta$ of each ablation against the full method; the absolute level of that row has no interpretation, since fitness is not comparable across algorithm families.

\begin{table}[!htbp]
\centering
\footnotesize
\caption{Per-task component ablation: audited L2 champion fitness, mean over 3 independent evolution runs per cell (evolution seeds 42/123/2026). Best per row in bold.}
\label{tab:ablation_full}
\setlength{\tabcolsep}{3pt}
\begin{tabular*}{\textwidth}{@{\extracolsep{\fill}}lcccccc}
\toprule
 & & \multicolumn{2}{c}{Ablating PCE} & \multicolumn{3}{c}{Ablating PPE} \\
\cmidrule(lr){3-4}\cmidrule(lr){5-7}
Task & Full & w/o decomp. & w/o Phase~1 & L0-only & L2-only & L1-only \\
\midrule
\multicolumn{7}{@{}l}{\emph{SAC}} \\
\quad dm-cheetah-run & \textbf{0.685} & 0.639 & 0.622 & 0.556 & 0.613 & 0.636 \\
\quad gym-HalfCheetah-v5 & \textbf{0.763} & 0.702 & 0.694 & 0.682 & 0.649 & 0.671 \\
\quad md-100envs & 0.907 & 0.945 & 0.954 & \textbf{0.957} & 0.754 & 0.728 \\
\quad mw-drawer-open-v3 & 0.780 & 0.763 & 0.763 & 0.718 & \textbf{0.782} & 0.730 \\
\quad\emph{SAC mean} & \textbf{0.784} & 0.762 & 0.758 & 0.728 & 0.700 & 0.691 \\
\addlinespace[2pt]
\multicolumn{7}{@{}l}{\emph{PPO}} \\
\quad gym-Hopper-v5 & 0.849 & 0.874 & 0.818 & 0.850 & \textbf{0.882} & 0.876 \\
\quad gym-HalfCheetah-v5$^{*}$ & 0.201 & 0.201 & 0.197 & 0.193 & \textbf{0.272} & 0.204 \\
\addlinespace[2pt]
\multicolumn{7}{@{}l}{\emph{DQN}} \\
\quad LunarLander-v3 & 0.752 & 0.465 & 0.638 & 0.600 & 0.689 & \textbf{0.772} \\
\quad MountainCar-v0$^{*}$ & 0.908 & 0.909 & 0.948 & 0.942 & \textbf{0.959} & 0.938 \\
\midrule
\quad\emph{Mean over all 8 pairs} & \textbf{0.731} & 0.687 & 0.704 & 0.687 & 0.700 & 0.694 \\
\bottomrule
\end{tabular*}
\end{table}

\section{Per-Cell Evolution Variance of the SAC Ablation}
\label{app:ablation_spread}

Table~\ref{tab:ablation_spread} restates the SAC block of Table~\ref{tab:ablation_full} with the per-cell standard deviation across the three independent evolution runs, together with the unevolved SAC baseline under the identical L2 protocol. These spreads set the resolution of the PCE comparison: the gaps being compared (0.02--0.04) are smaller than several of the cell standard deviations.

\begin{table}[!htbp]
\centering
\footnotesize
\caption{SAC component ablation with per-cell evolution variance: audited L2 champion fitness, mean $\pm$ std over \textbf{3 independent evolution runs} (evolution seeds 42/123/2026). The $\pm$ here is \emph{evolution} variance, not the \emph{training} variance reported in Table~\ref{tab:main_results}; the two are different quantities and should not be compared in magnitude. $\ddagger$~The baseline was never evolved, so no evolution variance exists for it; the spread on that row is its \emph{training} variance over the three seeds $\{0,42,123\}$ and is not comparable to the evolution variance above it. Best per column in bold.}
\label{tab:ablation_spread}
\setlength{\tabcolsep}{3pt}
\begin{tabular*}{\textwidth}{@{\extracolsep{\fill}}lccccc}
\toprule
Condition & dm-cheetah-run & HalfCheetah-v5 & md-100envs & mw-drawer-open & Avg \\
\midrule
Full method & \textbf{0.685} $\pm$ 0.025 & \textbf{0.763} $\pm$ 0.048 & 0.907 $\pm$ 0.095 & 0.780 $\pm$ 0.027 & \textbf{0.784} \\
\addlinespace[2pt]
\multicolumn{6}{@{}l}{\emph{Ablating PCE}} \\
\quad w/o decomposition & 0.639 $\pm$ 0.001 & 0.702 $\pm$ 0.083 & 0.945 $\pm$ 0.017 & 0.763 $\pm$ 0.051 & 0.762 \\
\quad w/o Phase~1 & 0.622 $\pm$ 0.018 & 0.694 $\pm$ 0.109 & 0.954 $\pm$ 0.011 & 0.763 $\pm$ 0.029 & 0.758 \\
\addlinespace[2pt]
\multicolumn{6}{@{}l}{\emph{Ablating PPE}} \\
\quad L0-only selection & 0.556 $\pm$ 0.059 & 0.682 $\pm$ 0.071 & \textbf{0.957} $\pm$ 0.044 & 0.718 $\pm$ 0.035 & 0.728 \\
\quad L2-only selection & 0.613 $\pm$ 0.007 & 0.649 $\pm$ 0.012 & 0.754 $\pm$ 0.136 & \textbf{0.782} $\pm$ 0.024 & 0.700 \\
\quad L1-only selection & 0.636 $\pm$ 0.019 & 0.671 $\pm$ 0.022 & 0.728 $\pm$ 0.150 & 0.730 $\pm$ 0.156 & 0.691 \\
\midrule
SAC baseline$^\ddagger$ & 0.302 $\pm$ 0.037 & 0.340 $\pm$ 0.014 & 0.677 $\pm$ 0.172 & 0.706 $\pm$ 0.265 & 0.506 \\
\bottomrule
\end{tabular*}
\end{table}

\section{DQN Component Ablation}
\label{app:ablation_dqn}

We repeated the component ablation of Section~\ref{sec:ablation} (full method vs.\ w/o Phase~1 vs.\ w/o component decomposition) on the two DQN tasks, 3 independent evolution runs per cell.

Only LunarLander is informative. MountainCar is saturated: all three conditions land within 0.908--0.948 (and within 0.91--0.96 across all six conditions once the PPE ablations are included), a spread comparable to the within-cell seed noise, so the task cannot separate the conditions---and there the no-Phase-1 condition is nominally \emph{ahead} of the full method. We report the MountainCar column for completeness but draw no conclusion from it, and we do not average the two tasks together.

On LunarLander the SAC ordering holds directionally---full 0.752 $>$ w/o Phase~1 0.638 $>$ w/o decomposition 0.465, a $+$0.287 margin over the monolithic condition. With $n{=}3$ runs per condition, however, the smallest attainable one-sided Wilcoxon $p$ is 0.125, so significance is out of reach by construction; this is directional replication only.

\begin{table}[!htbp]
\centering
\small
\caption{DQN component ablation (3 independent evolution runs per cell; mean $\pm$ std of the audited L2 champion fitness). LunarLander replicates the SAC ordering full $>$ w/o Phase~1 $>$ w/o decomposition directionally. MountainCar is saturated---all conditions within 0.908--0.948---and is not read; no mean is taken across the two tasks. Best per row in bold.}
\label{tab:ablation_dqn}
\begin{tabular*}{0.85\textwidth}{@{\extracolsep{\fill}}lccc}
\toprule
Task & Full & w/o Phase~1 & w/o decomp. \\
\midrule
LunarLander-v3 & \textbf{0.752} $\pm$ 0.027 & 0.638 $\pm$ 0.064 & 0.465 $\pm$ 0.084 \\
MountainCar-v0 (saturated) & 0.908 $\pm$ 0.006 & \textbf{0.948} $\pm$ 0.009 & 0.909 $\pm$ 0.049 \\
\bottomrule
\end{tabular*}
\end{table}

\section{Component-Level Analysis}
\label{app:components}

Tables~\ref{tab:component_detail_a} and~\ref{tab:component_detail_b} describe the five component implementations of all 17 champion programs of Table~\ref{tab:main_results}. Each cell was written by reading the corresponding function in the champion's source file, not by paraphrasing the LLM's own commit message; descriptions name the mechanism that is actually executed on the SAC code path. All 17 programs additionally carry an unreachable on-policy (GAE/PPO) branch inherited from the shared function template; those branches are ignored here because they never execute under SAC.

\begin{table}[!htbp]
\centering
\scriptsize
\setlength{\tabcolsep}{3pt}
\caption{Component implementations of the champion programs, DeepMind Control and Gym locomotion (9 tasks). ``Standard SAC'' and ``unchanged'' mark slots that are \emph{not} meaningfully evolved away from the seed program; $p$ denotes training progress in $[0,1]$.}
\label{tab:component_detail_a}
\begin{tabular}{p{1.55cm}p{2.2cm}p{2.2cm}p{2.2cm}p{2.2cm}p{2.2cm}}
\toprule
Task & Encoding & Value Target & Critic Loss & Policy Loss & Exploration \\
\midrule
\multicolumn{6}{l}{\textit{DeepMind Control}} \\
dm-cheetah-run & $\mathrm{arcsinh}(x)$, no rescale & Fixed $0.95\min$ $+\,0.05\max$ & Huber ($\beta{=}1$) $+\,10^{-5}Q^2$ & Standard SAC $+\,10^{-3}$ mean L2 & $5\times$ reward, EMA curiosity, action L1 \\
dm-walker-walk & $5\,\mathrm{arcsinh}(x/5)$, clip $[-25,25]$ & Sine anneal mean$\to$min, then variance penalty & Asym.\ Huber, EMA-adaptive $\delta$, $\tau$ 0.55--0.75 & Cosine min/max blend; asym.\ $\alpha$ LR $5\times$/$1\times$ & Dual-timescale EMA novelty, cosine decay \\
dm-walker-run & $5\,\mathrm{arcsinh}(x/5)$ + decaying noise & Per-sample min/mean from Q-disagreement & Asym.\ Huber, quantile-EMA $\delta$ & Softmin Q ($\beta$ 2$\to$10), PI $\alpha$ controller & Convexified reward + movement synergy \\
dm-quadruped-walk & Clip $[-20,20]$, then symlog & Cosine-annealed max-Q weight $0.1\to0$ & Scaled log-cosh ($\beta{=}2$) $+\,10^{-5}Q^2$ & Standard SAC $+\,10^{-3}$ action L2 & $10\times$ reward + decaying movement bonus \\
\midrule
\multicolumn{6}{l}{\textit{Gym MuJoCo}} \\
gym-HalfCheetah-v5 & Clip $[-20,20]$ only & Min-weight anneal $0.5\to0.95$ & Standard Huber ($\beta{=}1$) & Standard SAC $+\,10^{-3}$ mean L2 & Welford z-score novelty, off at $p{=}0.6$ \\
gym-Hopper-v5 & Piecewise log above $|x|{>}5$, clip $[-20,20]$ & \textbf{Standard SAC} $\min(Q_1,Q_2)$ & Huber, fixed $\delta{=}10$ & \textbf{Standard SAC} & \textbf{None}---returns reward unchanged \\
gym-Walker2d-v5 & Symlog (scale 5) + noise over first 10\% & Softmin Q, temperature $1.0\to0.1$ & Asym.\ Charbonnier, $\tau$ $0.40\to0.60$ & Min$\to$mean blend + disagreement penalty & RFF-RND novelty, $e^{-4p}$ decay, action L2 \\
gym-Humanoid-v5 & Symlog + deadzone, clip $[-20,20]$ & Optimistic min/max blend, weight $\to 0$ & Huber, $\delta{=}$EMA$|$TD$|$, growing asymmetry & \textbf{Standard SAC} & Random-projection Mahalanobis novelty \\
gym-Ant-v5 & Clip $[-20,20]$ only & \textbf{Standard SAC} $\min(Q_1,Q_2)$ & \textbf{Plain MSE (unchanged)} & Standard SAC $+\,10^{-3}$ mean L2 & $0.05\log$ of EMA-scaled $\|\Delta\mathrm{obs}\|$, linear decay \\
\bottomrule
\end{tabular}
\end{table}

\begin{table}[!htbp]
\centering
\scriptsize
\setlength{\tabcolsep}{3pt}
\caption{Component implementations of the champion programs, Meta-World manipulation and MetaDrive driving (8 tasks). On these benchmarks the encoder is where most of the evolved content sits: every champion rewrites the raw observation vector into task-specific geometric features.}
\label{tab:component_detail_b}
\begin{tabular}{p{1.55cm}p{2.2cm}p{2.2cm}p{2.2cm}p{2.2cm}p{2.2cm}}
\toprule
Task & Encoding & Value Target & Critic Loss & Policy Loss & Exploration \\
\midrule
\multicolumn{6}{l}{\textit{Meta-World}} \\
mw-push-v3 & Unused dims $\to$ distance/velocity features, clip $[-10,10]$ & Softmin blend, $\kappa$ $5.0\to0.5$, $w\le0.4$ & Asym.\ Huber, EMA $\delta$, $\tau$ $0.8\to0.5$ & Batch-normalized Q + UCB disagreement bonus & Potential shaping ($c{=}50/200$) + EMA curiosity \\
mw-drawer-open-v3 & Inject distance/contact/alignment scalars, then symlog & Variance-aware min/mean, pessimism $1.0\to0.2$ & Quantile-EMA Huber (0.85), TD reweighting & Min-Q $-$ decaying disagreement; action penalties & Adaptive potential shaping, decay floor 0.2 \\
mw-door-open-v3 & Velocity rebuild + proximity features, scaled symlog & Min-Q $+\,0.25$ cosine-decayed optimism & Log-cosh Huber, EMA $\delta$, $+\,10^{-4}Q^2$ & $0.95\min$ $+\,0.05\max$; 0.05 disagreement penalty & Potential shaping + object motion + grasp term \\
mw-peg-insert-side-v3 & History dims $\to$ velocity/direction features & Min-Q $+\,\beta(\max{-}\min)$, $\beta$ $0.15\to0$ & Standard Huber, MAD-derived EMA $\delta$ & \textbf{Standard SAC} (+ inactive saturation hinge) & Potential shaping (inverse/exp/grasp), $0.30\to0.05$ \\
mw-pick-place-v3 & Overwrite dims 18--29 with TCP/object/goal vectors & Min-Q $+\le15\%$ disagreement-gated optimism & Huber, $\delta{=}$EMA 80th-pct TD error & Standard SAC + mean L2 annealed to 0 & Potential shaping on reach/place/$z$ distances \\
\midrule
\multicolumn{6}{l}{\textit{MetaDrive}} \\
md-validation & Symlog ego; LiDAR $\to$ forward-weighted sector statistics & $0.9\min+0.1\max$, target clamp $[-50,2000]$ & Expectile-weighted Huber ($\beta{=}2$) & Boundary penalty + per-channel action regularizers & Symlog reward + EMA ego novelty, off at $p{=}0.5$ \\
md-10env & EMA z-score ego + tanh clip; LiDAR 3-beam max-pool & Mean $-$ trust-gated disagreement; $\gamma$ $0.90\to1.00$ & Asym.\ log-cosh, EMA $\kappa\in[1,5]$ & Mean $-$ pessimism$\,\cdot\,$std; PI $\alpha$ controller & LiDAR safety margin + novelty + stagnation penalty \\
md-100envs & Symlog ego; LiDAR avg/min smoothing $\to(1{-}d)^2$ & Pessimism $0.5{+}0.5p$ blend of min and mean & Focal-weighted adaptive-$\delta$ Huber & Advantage-weighted BC trust region + boundary penalty & LiDAR approach-rate penalty + throttle curriculum \\
\bottomrule
\end{tabular}
\end{table}

Read across all $17\times5=85$ slots, the tables support a more restrained reading of the method than the per-task highlights of Appendix~\ref{app:innovations} do, and we state the three caveats explicitly.

\textbf{Not every component slot is meaningfully evolved.} Three of gym-Ant-v5's five components are the unmodified seed program---a clip-only encoder, a textbook $\min(Q_1,Q_2)$ target, and plain MSE critic loss---and gym-Hopper-v5's exploration component consists of \texttt{return float(reward)}, i.e.\ the slot is switched off entirely. Four champions (Hopper, Humanoid, peg-insert-side, pick-place) retain the standard SAC actor loss up to a small action regularizer. Phase~1 spends its bandit pulls on all five arms regardless, so an evolved program is not a program in which all five components changed; it is one in which the components that mattered changed. This is consistent with the ablation result of Section~\ref{sec:ablation}, where the margin of per-component mutation over rewriting the whole program is smaller than the run-to-run spread of the cells being compared.

\textbf{Several annealed mechanisms converge back to the baseline.} dm-quadruped-walk's max-Q weight decays $0.1\to0$, gym-Humanoid-v5's optimism weight decays to $0$, and gym-Walker2d-v5's softmin temperature decays to a hard minimum. In each case the champion is a deviation from clipped double-Q early in training and is exactly clipped double-Q at convergence.

\subsection{The PPO and DQN Champions}
\label{app:components_xalgo}

Tables~\ref{tab:component_detail_ppo} and~\ref{tab:component_detail_dqn} give the same slot-by-slot reading for the 5 PPO and 4 DQN champions of Table~\ref{tab:main_results}. As above, every cell describes the mechanism that actually \emph{executes} under that family's trainer rather than the mutation's own rationale comment, and ``unchanged'' is measured against the vanilla seed program (identity encoder, textbook GAE / one-step Bellman target, plain MSE, textbook clipped surrogate, unshaped reward), not against the champion's own run directory---each run's local starting program is already a Phase-1 evolved program, so diffing against it would systematically understate what evolution changed.

\paragraph{Which slots are live under each trainer.} This determines what the counts below can mean. Our PPO trainer passes the on-policy value array and $\lambda$ into the value-target component and never passes twin-$Q$ arguments, so the off-policy branch every champion inherits from the shared template is unreachable---the mirror of the unreachable GAE branch in the SAC champions. Our DQN trainer calls only \emph{four} of the five components: \texttt{compute\_policy\_loss} is never imported, and the value-target and critic-loss components are called with the twin-$Q$ arguments, the entropy coefficient and the second critic's predictions all set to null, which makes the GAE branch, the twin-$Q$ aggregation branch, the entropy-bonus term and the second-critic term dead in all four DQN champions. Crucially, the two mechanisms a reader would most expect a DQN champion to innovate on---the $\epsilon$-greedy schedule and the target-network synchronisation---are hardcoded harness hyperparameters \emph{outside all five slots}, so DQN evolution could never reach them. This is a property of our decomposition as implemented, not a discovery failure, and it is the single most important constraint on the cross-family comparison.

\begin{table}[!htbp]
\centering
\scriptsize
\setlength{\tabcolsep}{3pt}
\caption{Component implementations of the 5 PPO champion programs. $p$ denotes training progress in $[0,1]$. Every champion inherits an unreachable off-policy (twin-$Q$) branch from the shared template, which is ignored here. $^\mathsection$~Entropy term with a sign \emph{opposite} to our trainer's convention: see the caveats below.}
\label{tab:component_detail_ppo}
\begin{tabular}{p{1.55cm}p{2.2cm}p{2.2cm}p{2.2cm}p{2.2cm}p{2.2cm}}
\toprule
Task & Encoding & Value Target & Critic Loss & Policy Loss & Exploration \\
\midrule
gym-Hopper-v5 & Piecewise symlog above $|x|{>}5$, clip $[-15,15]$ & Cosine-annealed GAE $\lambda$ $0.98\to0.90$ + MC-return blend (weight $0.1p$) & Smooth-L1, $\beta$ $1.0\to0.2$, $+\,10^{-4}V^2$ & Dual clipping ($c{=}3$) + multiplicatively adapted KL coef.\ (target 0.015) + decaying entropy$^\mathsection$ & EMA-normalized state-difference novelty $-$ action L2 + survival bonus \\
gym-HalfCheetah-v5 & Symlog $\mathrm{sign}(x)\log(1{+}|x|)$ & Linear GAE $\lambda$ $0.98\to0.90$ & Huber ($\beta{=}1$) $+$ decaying $5{\times}10^{-5}V^2$ & Clip anneal $0.2\to0.1$, adv.\ clamp $\pm5$, KL penalty $0.02(1{-}p)$, entropy & $-0.005\,\mathbb{E}[a^2]$ + decaying state-magnitude bonus \\
gym-Ant-v5 & $5\tanh(x/5)$ + decaying Gaussian noise & Linear GAE $\lambda$ $0.95\to0.85$ + advantage mean-centering & Adaptive Huber, $\beta$ $2.0\to0.5$, $+\,10^{-4}V^2$ & Dual clipping ($c{=}1.5$) + KL penalty 0.005 + entropy $0.01e^{-3p}$ & Dimension-invariant action penalty + decaying survival bonus \\
gym-Walker2d-v5 & Symmetric clip $[-10,10]$ & Linear GAE $\lambda$ $0.95\to0.90$ & Adaptive-$\delta$ Huber, $\delta{=}$ batch median $|$TD$|$, clamp $[0.1,10]$ & Dual clipping ($c{=}3$) + hinge KL penalty $\mathrm{relu}(\mathrm{KL}{-}0.015)$, entropy & Decaying action-energy penalty $-0.005\Sigma a^2(1{-}p)$ \\
gym-Humanoid-v5 & Piecewise log above $|x|{>}2$, linear below & \textbf{Dual-$\lambda$ GAE}: parallel $\lambda{=}0.97$ and $0.90$ streams, averaged 50/50 & Huber, fixed $\beta{=}10$ (MSE-like in range, linear on outliers) & Clip anneal $0.2\to0.05$ + dual clipping + KL-mask pseudo early-stopping + KL penalty & Decaying L1 \emph{and} L2 action penalties, clip $\pm10$ \\
\bottomrule
\end{tabular}
\end{table}

\begin{table}[!htbp]
\centering
\scriptsize
\setlength{\tabcolsep}{3pt}
\caption{Component implementations of the 4 DQN champion programs. Bracketed \emph{[dead]} annotations name mechanisms present in the file but unreachable under our DQN trainer; they are excluded from all counts. The policy-loss slot is never called at all, so no DQN champion can innovate there.}
\label{tab:component_detail_dqn}
\begin{tabular}{p{1.55cm}p{2.2cm}p{2.3cm}p{2.2cm}p{1.7cm}p{2.2cm}}
\toprule
Task & Encoding & Value Target & Critic Loss & Policy Loss & Exploration \\
\midrule
CartPole-v1 & Welford online standardization $\to$ symlog $\to$ clip $\pm20$ & \textbf{Unchanged} $r{+}\gamma(1{-}d)\max_a Q'$ \emph{[dead: $0.8\min{+}0.2\max$ aggregation]} & \textbf{$0.5\times$MSE---unchanged} up to a constant \emph{[dead: 2nd-critic term]} & \textbf{Never called} & Welford-normalized transition distance, positive part, $0.01(1{-}p)$ \\
MountainCar-v0 & Symlog $\to$ EMA mean/var ($\alpha{=}0.005$) $\to$ clip $\pm5$ & \textbf{Unchanged} \emph{[dead: uncertainty-weighted min/max mixing]} & Adaptive expectile regression, weights $0.7/0.3\to0.5/0.5$ & \textbf{Never called} & Online RND with random Fourier features, $0.2\,\mathrm{relu}(z)(1{-}p)$ \\
LunarLander-v3 & Symmetric clip $[-20,20]$ & \textbf{Unchanged} \emph{[dead: adaptive min/mean blend \emph{and} the advertised adaptive-$\gamma$ warm-up]} & Smooth-L1 ($\beta{=}1$)---a genuine live change from seed MSE & \textbf{Never called} & State-difference bonus (L2, cap 5.0), first 20\% of training only, linearly decayed \\
Acrobot-v1 & Clip $\pm10$ + negligible decaying noise & \textbf{Unchanged} + live reward clamp $\pm1000$ (inert for Acrobot) \emph{[dead: optimistic min/max mixing]} & Asymmetric Huber, under/over weights $1.2/0.8$, $+\,10^{-5}Q^2$ & \textbf{Never called} & Online RND (RFF) + NLMS predictor, EMA novelty, $e^{-4p}$ decay \\
\bottomrule
\end{tabular}
\end{table}

\paragraph{PPO innovates in the slots that are characteristically PPO, in every champion.} All 5 PPO champions modified all five slots. More informative than the count is \emph{where} the content sits: 5/5 value-target modifications are GAE-specific ($\lambda$ scheduling in 4/5---three linear, one cosine---dual-$\lambda$ streams in 1/5, Monte-Carlo blending in 1/5, advantage centering in 1/5), and no champion touched the discount or the bootstrap structure. All 5 policy-loss modifications are clipped-surrogate-specific: dual clipping 5/5 and KL regularisation 5/5, the latter in four distinct forms (fixed coefficient, hinge, adaptive multiplicative, and mask-based pseudo early-stopping), plus clip-range annealing 4/5 and ratio or advantage clamping 4/5. These mechanisms have no SAC or DQN analogue, and the SAC signature of $Q$-annealing has no PPO analogue because PPO has no twin critics.

\paragraph{DQN's live surface is three slots, and it uses all three.} Counting only reachable code, the DQN champions modified encoding 4/4, exploration 4/4 and the critic loss 3/4 (asymmetric Huber on Acrobot, expectile regression on MountainCar, Huber on LunarLander; CartPole's $0.5\times$MSE is a constant rescaling absorbable into the learning rate and we score it unchanged). The live value target is the seed $\max_a Q'$ backup in 4/4, and the policy slot is 0/4 by construction. Two independent implementations of RND with random Fourier features arose on MountainCar and Acrobot, which is the clearest evidence that evolution used the room it did have. Counting dead code would have turned the value-target column from 0/4 into 4/4, so we state the live counts and not the nominal ones.

\paragraph{One consensus innovation survives across all three families, and it is the critic loss.} Replacing MSE with a robust regression loss is reached independently under all three trainers: it is the SAC signature (16/17), it appears in 5/5 PPO champions (adaptive $\beta$ in 3, median-derived adaptive $\delta$ in 1, fixed $\beta{=}10$ in 1) and in 3/4 DQN champions---\textbf{8 of the 9 PPO and DQN champions}, under two different trainers. A second, weaker consensus is symlog/clip-family observation encoding (SAC, 5/5 PPO, 4/4 DQN), but this slot's seed is a bare identity function, so nearly any change registers and the column is the least discriminating of the five. Beyond these two, the families are largely disjoint, and the disjointness is the informative part: each family's other innovations sit in the machinery specific to that family.

\paragraph{What this does and does not establish.} The strong reading---that each family innovates in the slots characteristic of \emph{that} family---holds cleanly for PPO and does not hold for DQN. But the DQN shortfall is structural rather than empirical: the slots that would carry a characteristic DQN innovation are not exposed by the decomposition as implemented, and the value-target slot that \emph{is} exposed degenerates to a fixed $\max_a Q'$ backup. The defensible claim is therefore that the decomposition supports genuine, family-appropriate innovation where the family has expressive room in the slots (SAC and PPO: five slots each) and degenerates where it does not (DQN: three live slots). We do \emph{not} claim the three families have equal searchable surface area; DQN is materially narrower.

\paragraph{Caveats specific to the cross-algorithm champions.} Three are worth stating plainly. \textbf{(1)~Dead off-policy branches.} All four DQN champions carry unreachable twin-$Q$ aggregation, GAE, entropy and second-critic terms, and three of the four rationale comments headline their $Q$-aggregation as the main change. \textbf{(2)~LunarLander's advertised adaptive-$\gamma$ warm-up is dead}, sitting inside the never-executed twin-$Q$ branch; the live target uses plain $\gamma$. This is the most tempting mis-attribution in the set. \textbf{(3)~One PPO champion carries an entropy term with a sign opposite to the trainer's.} Our trainer's own convention contributes $+\mathrm{coef}\cdot\overline{\log \pi}$ to the loss, which maximises entropy. The gym-Hopper-v5 champion instead adds $-\mathrm{coef}\cdot\overline{\log \pi}$, an entropy \emph{penalty} rather than a bonus, and still beats its baseline. The gym-HalfCheetah-v5 and gym-Walker2d-v5 champions write the bonus through double negatives but are algebraically identical to the trainer's convention; we flag only the Hopper champion, in Table~\ref{tab:component_detail_ppo}. Finally, these are the nine best-run champions only; we make no claim that these component choices recur across the other evolution seeds of the same task.

\end{document}